\documentclass[10pt, a4paper, twocolumn, teaser, showabstract]{naverlabseurope}

\usepackage{multirow}
\usepackage{multicol}
\usepackage{rotating}
\usepackage{bbm}
\usepackage{algorithm2e,algpseudocode}
\usepackage{hyperref}
\usepackage[capitalise]{cleveref}
\usepackage{url}

\newlength{\smallimage}
        \definecolor{rel}{rgb}{.1,.6,.2}
        \definecolor{nrl}{rgb}{1,1,1}
        \definecolor{qim}{rgb}{1,1,1}

\def\eg{\emph{e.g.}}

\def\ie{\emph{i.e.}}

\def\cf{\emph{c.f.\,}}

\def\vs{\emph{versus\,\,}}
\def\wrt{w.r.t.\,}

\definecolor{lightgray}{gray}{0.93}

\def\be{\begin{equation}}
\def\ee{\end{equation}}
\def\bea{\begin{eqnarray}}
\def\eea{\end{eqnarray}}
\def\ben{\begin{eqnarray*}}
\def\een{\end{eqnarray*}}

\def\bi{\begin{itemize}}
\def\ei{\end{itemize}}

\newcommand{\btab}[1]{\begin{tabular}{#1}}
\newcommand{\etab}{\end{tabular}}
\newcommand{\ba}[1]{\begin{array}{#1}}
\newcommand{\ea}{\end{array}}

\DeclareMathOperator*{\argmax}{\mathrm{argmax}}

\def\Re{{\rm I\!R}}                            
\def\<{\langle}
\def\>{\rangle}



\definecolor{DarkGreen}{rgb}{0.5, 0.9, 0.5}

\newcommand{\bd}[1]{{#1}}

\newcommand{\PPNeSF}{{ppNeSF}}
\newcommand{\ZipNeRFwoRGB}{{ZipNeRF-wo-RGB}}
\newcommand{\RGBPPNeSF}{{RGB-ppNeSF}}
\newcommand{\NIF}{NIF}
\newcommand{\NIFs}{NIFs}
\newcommand{\SF}{${\bf{\Omega}}$}
\newcommand{\FF}{${\bf{\Gamma}}$}
\newcommand{\GF}{${\bf{\Psi}}$}
\newcommand{\IE}{${\bf{\Phi}}$}

\newcommand{\PAR}[1]{\vskip2pt \noindent{\bf #1.}}
\newcommand{\PARR}[1]{\vskip2pt \noindent{\bf #1}}

\def\softmax{\textrm{softmax}}
\def\mean{\textrm{mean}}

 \title{Can we make NeRF-based visual localization privacy-preserving?}

\authors{Maxime Pietrantoni$^{1,2,3}$ and 
\authsep 
Martin Humenberger$^3$ and
\authsep 
Gabriela Csurka$^3$ and
\authsep 
Torsten Sattler$^2$}

\affiliations{
$^1$ Faculty of Electrical Engineering, Czech Technical University in Prague, \url{\{firstname.lastname\}@cvut.cz}\\
$^2$ Czech Institute of Informatics, Robotics and Cybernetics, Czech Technical University in Prague \\
$^3$ NAVER LABS Europe, \url{\{firstname.lastname\}@naverlabs.com} \\ } 
\contributions{}
\websiteref{}
\website{}
\date{}

\teaserfig{\includegraphics[width=11cm,height=7.5cm]{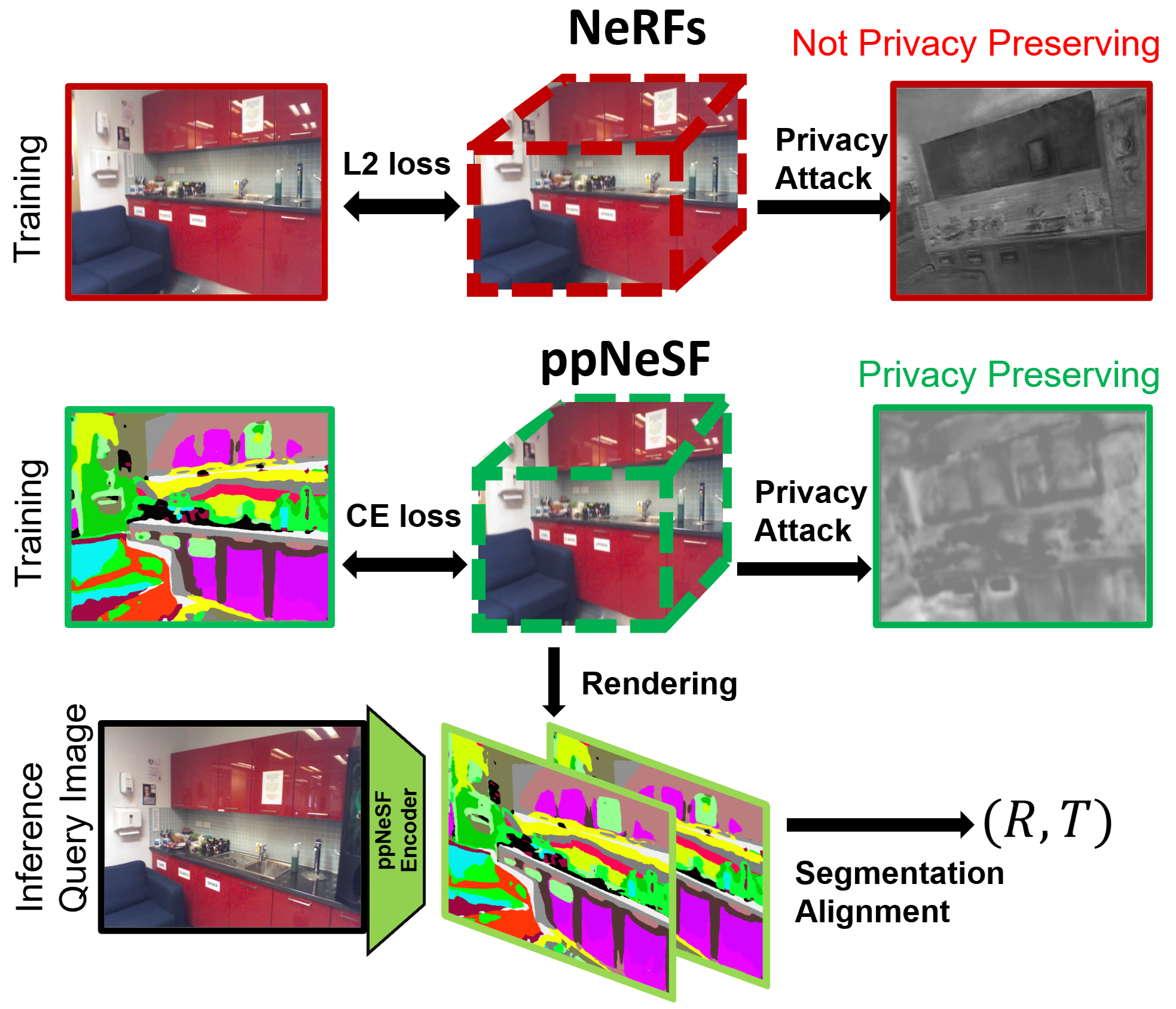}}
\teasercaption{NeRFs optimized with photometric losses 
store fine-grained scene details, making them susceptible to privacy attacks. We introduce ppNeSF, which replaces RGB supervision with segmentation-based training, ensuring that only high-level structural information is stored. 
\textbf{ppNeSF} enables accurate visual localization by aligning segmentation maps via 
pose refinement.   \label{fig:teaser}}
\date{}

\begin{abstract}
Visual localization (VL) is the task of estimating the camera pose in a known scene. VL methods, a.o., can be distinguished based on how they represent the scene, e.g., explicitly through a (sparse) point cloud or a collection of images or implicitly through the weights of a neural network. Recently, NeRF-based methods have become popular for VL. While NeRFs offer high-quality novel view synthesis, they inadvertently encode fine scene details, raising privacy concerns when deployed in cloud-based localization services as sensitive information could be recovered. In this paper, we tackle this challenge on two ends. We first propose a new protocol to assess privacy-preservation of NeRF-based representations. We show that NeRFs trained with photometric losses store fine-grained details in their geometry representations, making them vulnerable to privacy attacks, even if the head that predicts colors is removed. Second, we propose \textbf{ppNeSF} (Privacy-Preserving Neural Segmentation Field), a NeRF variant trained with segmentation supervision instead of RGB images. These segmentation labels are learned in a self-supervised manner, ensuring they are coarse enough to obscure identifiable scene details while remaining discriminativeness in 3D. The segmentation space of ppNeSF can be used for accurate visual localization, yielding state-of-the-art results.
\end{abstract}

\begin{document}
\maketitle

\section{Introduction}

Visual localization (VL), a core component of self-driving cars~\cite{HengICRA19ProjectAutoVisionLocalization3DAutonomousVehicle} and autonomous robots~\cite{LimIJRR15RealTimeMonocularImageBased6DoFLocalization}, refers to the task of estimating the 6DoF camera pose from which an image was captured. 
In essence, VL approaches compute the pose by aligning the image with a representation of the scene. 
Common types of representations include 
3D Structure-from-Motion (SfM) point clouds 
~\cite{SarlinCVPR19FromCoarsetoFineHierarchicalLocalization,HumenbergerIJCV22InvestigatingRoleImageRetrieval,SattlerICCV15HyperpointsFineVocabulariesLocRecogn,SarlinCVPR21BackToTheFeature,von2020gn,vonstumberg2020lmreloc,xu2020deep,lindenberger2021pixelperfect}, a database of images with known intrinsic and extrinsic parameters~\cite{ZhouICRA20ToLearnLocalizationFromEssentialMatrices,Zheng2015ICCV,Bhayani_2021_ICCV,Zhang06TDPVT,SattlerCVPR19UnderstandingLimitationsPoseRegression}, or the weights of neural networks
~\cite{KendallICCV15PoseNetCameraRelocalization,BrachmannCVPR17DSACLocalization,WalchICCV17ImagebasedLocalizationUsingLSTMs,WeyandECCV16PlanetPhotoGeolocationCNN,Shuai3DV21DirectPoseNetwithPhotometricConsistency,ShuaiECCV22DFNetEnhanceAPRDirectFeatureMatching,MoreauCORL22LENSLocalizationEnhancedByNeRFSynthesis,chen2023refinement,MoreauX23CROSSFIRECameraRelocImplicitRepresentation,zhou2024nerfect,pietrantoni2024jointnerf}.
Recently, methods leveraging neural radiance fields (NeRFs) \cite{mildenhall2020nerf} as the underlying 3D scene representation emerged. 
They solve VL by either matching or aligning features rendered from NeRFs 
\cite{chen2023refinement,MoreauX23CROSSFIRECameraRelocImplicitRepresentation,zhou2024nerfect,pietrantoni2024jointnerf}.
Contrarily to sparse SfM models, these methods benefit from the dense 3D nature of NeRFs and their ability to render consistent color, geometry, or additional information such as semantics or features. 

In practice, large-scale VL services are very likely to be deployed in the cloud. 
Hence, privacy is becoming a critical requirement.
Following~
\cite{pittaluga2019revealing,Chelani2021CVPRP,pietrantoni2023segloc,zhou2022geometry,wang2023dgc}, we define privacy as the inability to retrieve personally identifiable information (\eg, pictures, documents, or texture details) and assume that general scene details do not represent a privacy breach. Note that,  
preventing exposure of broad semantic/geometric information, \eg, the type of scene or the presence of object classes,  in the context of visual localization remains unlikely as a minimum level of discriminative information is required to localize with sufficient accuracy.

Early works quantified privacy based on image synthesis \cite{pittaluga2019revealing} or detection metrics \cite{pietrantoni2023segloc,Chelani3DV253ObfuscationBasedPPReprRecoverableUsingNN} of images recovered from scene representations, demonstrating potential privacy leaks.
We argue that using object detectors to assess privacy is suboptimal as they only detect a small predefined set of privacy sensitive classes and they can be unreliable.
To enhance existing evaluation protocols, we propose to characterize privacy through the descriptive capacity of advanced vision-language models (VLMs). These models can capture fine-grained details with comprehensive class granularity, hence better reflecting the privacy we seek to quantify than both closed- and open-set object detectors.

While privacy has been widely studied in the context of localization with SfM representations \cite{pittaluga2019revealing,DusmanuCVPR21PrivacyPreservingImageFeaturesViaAdversarialAffineSubspaceEmbeddings,pan2023privacy,Chelani2021CVPRP,SpecialeCVPR19PrivacyPreservingImageBasedLocalization,SpecialeICCV19PrivacyPreservingImageQueriesforCameraLocalization,pietrantoni2023segloc}, in this paper we are the first to consider it in the context of localization with neural implicit fields.\footnote{\cite{kong2023identity} also consider the privacy of NeRFs, using gradients rather than RGB values for face reconstruction. However, \cite{kong2023identity} do not provide any quantitative analysis for privacy-preservation.} 

As a first contribution, we 
design an inversion attack that reconstructs images by inverting rendered internal representations of such fields and
propose a novel evaluation protocol 
based on a strong vision-language model (LlaVa~\cite{liu2024visual})
to assess the degree of privacy preservation offered by neural implicit representations.
The proposed 
attack exposes the privacy liability of radiance fields optimized with RGB supervision by showing that fine texture details are stored in the part of the neural fields associated with raw geometry (which may be found in all implicit neural field-based localization methods~\cite{MoreauX23CROSSFIRECameraRelocImplicitRepresentation,chen2023refinement,pietrantoni2024jointnerf,zhou2024nerfect}). 
Thus, simply removing 
the colour branch
after training does not guarantee privacy. 

Therefore, as a second contribution, we propose our Neural Segmentation Field (\PPNeSF), the first privacy-preserving approach to visual localization based on neural fields. 
Intuitively, segmentation labels form non-injective mappings from RGB pixels to object instances encoding less detailed information than raw RGB images \cite{pietrantoni2023segloc,pietrantoni2025gaussian}. 
Thus, we replace the photometric supervision with segmentation label supervision. 
We derive these supervisory targets in a self-supervised manner by applying an optimal transport \cite{distances2013lightspeed} labeling procedure within a joint 2D/3D hierarchical feature embedding space, which ensures their robustness, viewpoint consistency, and local discriminativeness.
These segmentation labels not only guide geometric refinement of the neural field but also establish a unified segmentation space aligning 3D scene representations with a 2D image encoder. 
This enables localization through the alignment of segmentation maps (\cf \cref{fig:teaser}).

In summary, our contributions are: 
1) In \cref{sec:privacy}, we analyse the vulnerability of neural field-based models in terms of how much privacy-sensitive information can be retrieved. We propose a new protocol to assess this systematically.
2) In \cref{sec:ppnefs}, we present our privacy-preserving Neural Segmentation Field (\PPNeSF), which is the first solution to reconcile implicit neural fields with privacy-preserving representations by introducing a self-supervised training pipeline leveraging segmentation labels as the primary supervision. 
3) 
We show that our 
\PPNeSF{} model enables 
effective visual localization while providing a high level of privacy.

\begin{table*}[tt]
\resizebox{\linewidth}{!}{
  \begin{tabular}{cl||c|c|c|c|c|c|c|c}
   & &  \multicolumn{8}{c}{LPIPS($\uparrow$)/ FID ($\uparrow$) / Captions similarity ($\downarrow$)} \\
   &  Model  & Chess & Fire & Heads & Office & Pumpkin & Redkitchen & Stairs  & Average\\
    \hline
   \multirow{2}*{\begin{sideways} mip360 \end{sideways}}  
    & \ZipNeRFwoRGB &  0.53/233/0.76 & 0.55/347/0.65 & 0.56/323/0.61 & 0.55/230/0.66 & 0.54/217/0.70 & 0.57/232/0.65 & 0.56/173/0.77 &  0.55/250/0.68\\ 
   & \PPNeSF &  \bd{0.60}/\bd{313}/\bd{0.64} & \bd{0.62}/\bd{425}/\bd{0.41} & \bd{0.60}/\bd{389}/\bd{0.44} & \bd{0.62}/\bd{282}/\bd{0.53} & \bd{0.58}/\bd{284}/\bd{0.43} & \bd{0.60}/\bd{304}/\bd{0.34} & \bd{0.57}/\bd{257}/\bd{0.63} & \bd{0.59}/\bd{322}/\bd{0.4}\\
   \hline
      \hline
    &    & Bicycle & Bonsai & Counter & Garden & Kitchen & Room & Stump  & Average\\
\hline
 \multirow{2}*{\begin{sideways} 7S \end{sideways}}  
 & \ZipNeRFwoRGB & 0.69/321/0.65 & 0.49/193/0.80 & 0.53/240/0.72 & 0.64/150/0.66 & 0.62/321/0.52 & 0.57/301/0.71 & 0.66/328/0.60 & 0.6/264/0.7\\
  & \PPNeSF &  \bd{0.81}/\bd{297}/\bd{0.37} & \bd{0.71}/\bd{446}/\bd{0.39} & \bd{0.72}/\bd{470}/\bd{0.30} & \bd{0.81}/\bd{308}/\bd{0.39} & \bd{0.74}/\bd{381}/\bd{0.34} & \bd{0.69}/\bd{386}/\bd{0.34} & \bd{0.81}/\bd{448}/\bd{0.43} & \bd{0.76}/\bd{390}/\bd{0.32} \\
    \hline
       \hline
   &    & scene1 & scene2a & scene3 & scene4a & scene5 & scene6 &   & Average\\
    \hline
     \multirow{2}*{\begin{sideways} 7S \end{sideways}}  
    & \ZipNeRFwoRGB & 0.51/262/0.51 & 0.53/372/0.73 & 0.52/334/0.47 & 0.55/267/0.54 & 0.50/253/0.52 & 0.53/302/0.52 & & 0.52/298/0.55 \\
    & \PPNeSF & \bd{0.56}/\bd{311}/\bd{0.41} & \bd{0.61}/\bd{364}/\bd{0.55} & \bd{0.57}/\bd{324}/\bd{0.38} & \bd{0.61}/\bd{334}/\bd{0.39} & \bd{0.56}/\bd{279}/\bd{0.43} & \bd{0.57}/\bd{316}/\bd{0.40} & & \bd{0.58}/\bd{321}/\bd{0.43} \\
    \hline
    \bottomrule
    \end{tabular}
}
\caption{We evaluate privacy-preservation through perception metrics and 
KeyBert  similarity of LLaVa~\cite{liu2024visual} descriptions (captions similarity between the original and the reconstructed image).
Lower reconstruction quality (\ie, higher LPIPS and FID values) and lower semantic embedding similarities imply better privacy.  
The left column indicates on which dataset the inversion model was trained.}
      \label{tab:recons_privacy}
      \vspace{-.5cm}
    \end{table*}

\section{Related work}
\label{sec:related}

\PAR{Privacy-preserving visual localization} 
Given a (sparse) set of features and their 2D positions, it is possible to recover the original image via a so-called inversion attack~\cite{pittaluga2019revealing}.  
By extension, detailed and recognizable images of a scene can be obtained from sparse 3D point clouds (where each point is associated with a local image feature) by reprojecting these 3D points with their desciptors into the image.
To prevent inversion attacks, visual localization approaches based on geometric obfuscation modify scene representations by replacing 2D or 3D positions with lines or planes or by permuting point coordinates~\cite{GeppertECCV20PrivacyPreservingSFM,GeppertCVPR21PrivacyPreservingLocalizationMappingUncal,GeppertCVPR22PrivacyPreservingPartialLoc,MoonCVPR24EfficientPPVL3DRayCloud,pan2023privacy,ShibuyaECCV20PrivacyPreservingVisualSLAM,lee2023paired}.
Yet, 
it is possible to (approximately) recover the underlying point positions 
\cite{ChelaniCVPR21HowPrivacyPreservingAreLineClouds,lee2023paired,Chelani3DV253ObfuscationBasedPPReprRecoverableUsingNN}, thus again enabling inversion attacks.
Another line of work uses 
features or representations that are harder to invert.
\cite{pietrantoni2023segloc,pietrantoni2025gaussian} demonstrates that replacing high-dimensional descriptors with segmentation labels effectively prevents such inversion. 
\cite{zhou2022geometry,wang2023dgc} 
perform matching based solely on geometry. 
In this work, we show how to enable privacy-preserving visual localization based on segmentations in the context of neural implicit scene representations.  

\PARR{Camera pose refinement-based visual localization} 
methods minimize, \wrt the pose, 
the difference between the query image and a rendering obtained by projecting the scene representation using the current pose estimate~\cite{alismail2017photometric,engel2014lsd,EngelPAMI17DirectSparseOdometry,schopscvpr19DADSLAM,SchopsCVPR17MultiViewStereo}. The pose is iteratively refined from an initial pose estimate.
Similarly, deep features may be leveraged by minimizing the differences between features extracted from the image and projections of features stored in the scene~\cite{SarlinCVPR21BackToTheFeature,von2020gn,vonstumberg2020lmreloc,xu2020deep,lindenberger2021pixelperfect}.
Alternative to SfM models such as meshes, NeRfs or 
3DGS  \cite{GermainCWPRWS21FeatureQueryNetworks,liu2024gsloc,sidorov2024gsplatloc,trivigno2024unreasonable} may be used to store the features. 
In the context of privacy-preserving VL, \cite{pietrantoni2023segloc,pietrantoni2025gaussian} use pose refinement to align segmentation maps. 
Our work uses a similar strategy for VL. 

\PAR{Visual localization using neural implicit fields} 
Leveraging the capability of novel view synthesis of NeRFs, iNeRF~\cite{yenchen2021inerf,lin2023parallel} proposes to perform pose refinement through photometric alignment, where the
optimization is performed by 
back-propagation through the neural field. 
Instead of direct photometric alignment, pose estimation can be performed by feature-metric alignment or by feature matching followed by PnP,  
either by training a feature field to replicate features provided by a pre-trained 2D encoder~\cite{chen2023refinement,MoreauX23CROSSFIRECameraRelocImplicitRepresentation,huang2025sparse} or 
by jointly training a 2D encoder
with the feature field in an unsupervised manner~\cite{pietrantoni2024jointnerf,zhou2024nerfect,pietrantoni2025gaussian}.
In contrast to feature alignment, we perform pose estimation through hierarchical alignment of 2D segmentation maps and 3D segmentation maps rendered from our privacy-preserving \PPNeSF{}  
model.
The most similar line of work is SegLoc~\cite{pietrantoni2023segloc} and GSFF~\cite{pietrantoni2025gaussian}, however, both approaches rely on 
explicit 3D primitives in the form of points or Gaussians. Instead, our method is based on a dense implicit neural field, optimized directly with self-supervised segmentation labels, leading to higher localization performances.

\section{Privacy vulnerability of NeRFs}
\label{sec:privacy}
Following prior work~
\cite{pietrantoni2023segloc,wang2023dgc,zhou2022geometry}, we do not consider raw coarse geometry as a privacy leak. 
We define privacy as the inability to retrieve sensitive information from a scene such as textures, text, or fine details that could reveal personal details. 
Following this definition, this section introduces 
an inversion attack and evaluation protocol to assess the degree of privacy of neural implicit fields (\NIFs). 
For any 3D point in a scene, a \NIF{} predicts a volumetric density which can be rendered to obtain depth maps. By extension, any other quantity (such as color,semantics,features) may be predicted through a second MLP (conditioned on the first one, creating another implicit field) and may also be rendered.
NeRFs are a type of \NIFs{} containing a color/radiance field. By opposition, our \PPNeSF{} is a \NIF{} but not a NeRF as it contains only geometric and segmentation fields.

Our main insight is that 
optimizing NeRFs with a photometric loss leads to texture and fine information being embedded in the geometric part of the model. 
Hence, the naive approach of 
simply removing the color prediction head after training does not make NeRFs privacy preserving. 
\cref{sec:ppnefs} later introduces our \PPNeSF{} solution to remedy to this issue.

In this section, we solely focus on evaluating the privacy of \NIFs{} as the main novelty and show that \PPNeSF{} is a privacy-preserving \NIF.
Privacy experiments  
for segmentations maps used in \PPNeSF{}
are provided in \cref{sec:segmprivacy}.

\begin{figure*}[ttt]
\centering
\includegraphics[width=16.5cm,height=3.3cm]{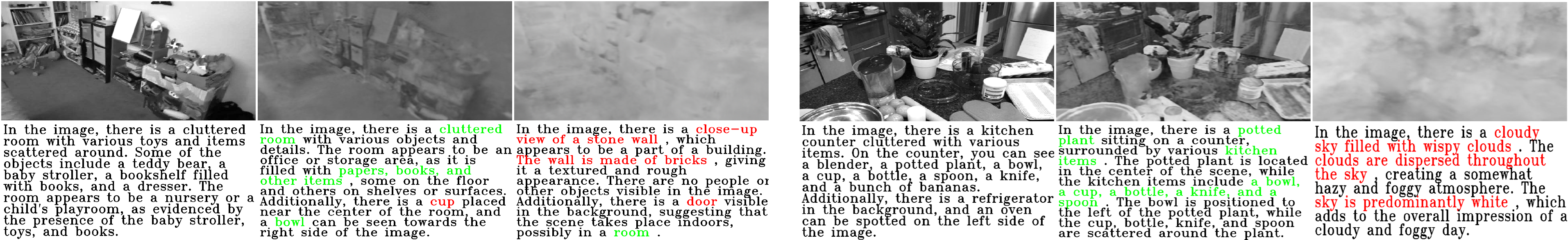}
\caption{Left to right: ground truth image and images reconstructed from \ZipNeRFwoRGB{} and \PPNeSF{} via our inversion attack. We also show 
the associated LlaVa's descriptions, highlighting objects correctly identified by the VLM in {\color{green}green}  and hallucinated objects 
in {\color{red}red}. 
The reconstructions from \PPNeSF{} do not contain any textural or fine-grained information, yielding completely wrong LlaVa's descriptions. In contrast, images reconstructed from \ZipNeRFwoRGB{} show 
much more details and most objects 
can be correctly identified by LlaVa.} 
\label{fig:visu_recons}
\end{figure*}

\PAR{Privacy attack}
To assess the degree of privacy of a 
\NIF, we want to quantify the amount of sensitive information it contains in its internal representation. To that end, we derive the following privacy attack.
 We call "internal representation" the feature output of the MLP predicting geometric information. In NeRFs this feature further serves as the input to the MLPs predicting color.
We choose this feature output as it is a component shared among many NeRF architectures \cite{mildenhall2020nerf,barron2022mip,wang2023neus,barron2023zip}) However, any other feature component in the geometry branch could have been used.
We aim at extracting the information contained in these internal representations. 
We thus train an inversion model\footnote{Note that this inversion model is independent from the proposed \PPNeSF{} solution and is only used to evaluate the privacy of \NIF{} models.}
that takes the rendered internal representation as input and reconstructs a grayscale image corresponding to the viewpoint from which the features were rendered. 
We learn to reconstruct  grayscale images instead of RGB images to increase the generalization power of the model across datasets and to make the model robust to color variations of certain objects. 
The inversion model is trained on scenes from one dataset and evaluated on scenes from another unseen one. 
Training details are provided in the \cref{sec:training}.

\PAR{Privacy evaluation protocol}
We now evaluate the quality of the images obtained through the privacy attack. If images are faithfully reconstructed, the \NIF{} model contains too much sensitive information and is not privacy preserving (and inversely). 
Given a \NIF{} optimized for a target scene 
and an inversion network (trained on a different set of scenes), we render the internal representation from each training image in the target scene. 
They are fed to the inversion model, which reconstructs a grayscale image associated with each viewpoint.

To measure the quality of the reconstructed images, we use 
two complementary measures as proxies for assessing the privacy of the \NIF:
\noindent i) The first 
evaluates the \textbf{visual quality} of the reconstructed image using standard 
perceptual metrics, such as LPIPS~\cite{zhang2018unreasonable} and FID~\cite{heusel2017gans}.
\noindent ii) The second 
evaluates how much \textbf{semantic information} can be recovered from the reconstructed image. 
While prior works use object detectors, 
we propose to use a strong visual-language model~\cite{liu2024visual}, 
which describes the content of the images in finer details than an object  
detector would.
We evaluate the semantic similarity between the descriptions of the reconstructed and the original image. 
Concretely, we encode each description using KeyBert~\cite{grootendorst2020keybert} 
and compute the cosine similarities between the resulting two description embeddings.

\PAR{Qualitative and quantitative results}
We evaluate the degree of privacy of our proposed model \PPNeSF{} (described in \cref{sec:ppnefs}) against ZipNeRF \cite{barron2023zip}, where the RGB head is removed after optimization (we call this privacy baseline \ZipNeRFwoRGB{}). We pick ZipNeRF as a comparison baseline for its SOTA novel view synthesis capabilities, but any other model could be considered.
We optimize \ZipNeRFwoRGB{} and \PPNeSF{} for each scene of the  7Scenes~\cite{ShottonCVPR13SceneCoordinateRegression}, mip360v2~\cite{barron2022mip}, and Indoor6~\cite{do2022learning} datasets. 
We train one inversion model per dataset (and per baseline) and evaluate it on the other datasets. 
\cref{tab:recons_privacy} shows results.

\begin{figure*}[tt]
\centering\includegraphics[width=0.90\linewidth]{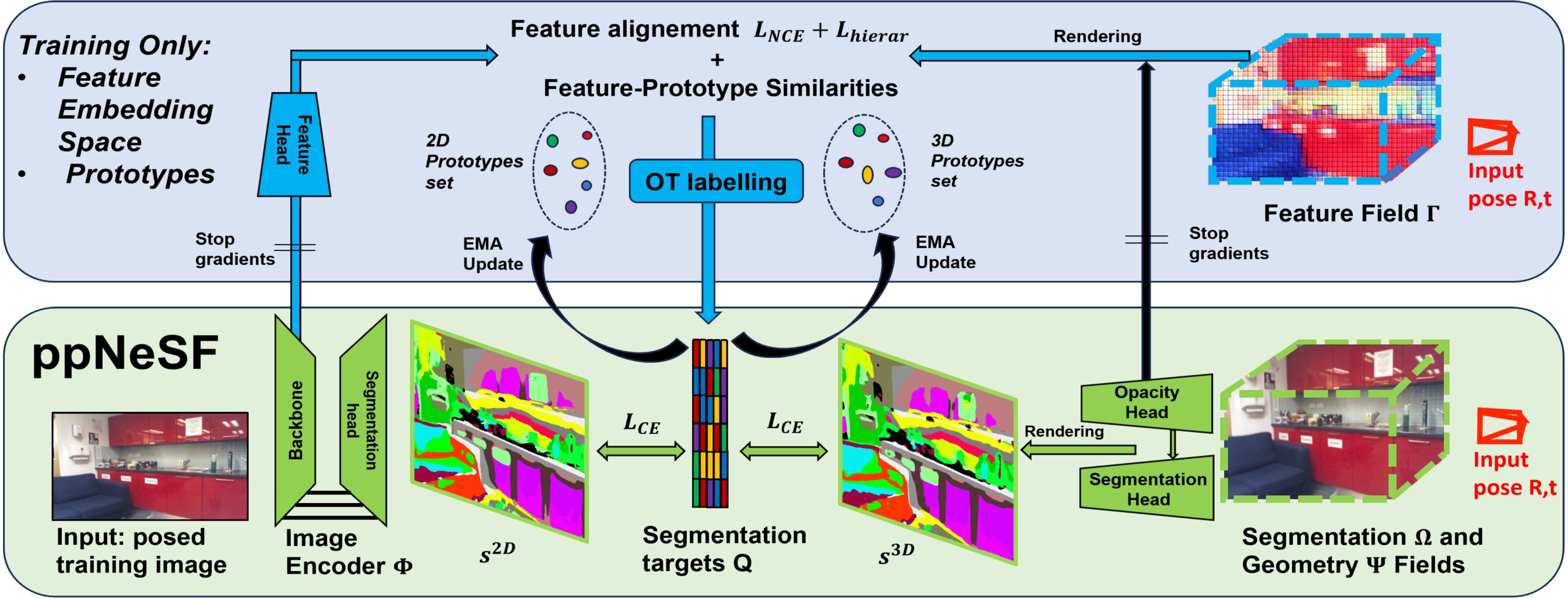} 
\caption{\textbf{The \PPNeSF{} architecture.} Given an input image and its associated input pose ($R$, $t$), we extract image-based features and segmentations from the Image Encoder \IE{}, render volumetric features from the feature field \FF{} and segmentations from \SF{}. The 2D and 3D features are aligned through hierarchical contrastive losses (top). Segmentation targets Q are derived through optimal transport based on feature/prototype similarities. The Encoder \IE{}, the Geometric \GF{} and Segmentation \SF{} Fields are optimized through a cross entropy loss against the targets (bottom). Finally the prototypes are updated with an Exponential Moving Average (EMA) scheme.
}
\label{fig:pipeline}
\end{figure*}

As \NIFs{} are optimized independently per scene and nothing constrains the internal representations of \NIFs{} trained on different scenes (for the same architecture) to lie in a common embedding space.
As such, there are no guarantees that the inversion model can generalize to \NIFs{} trained on scenes unseen during training.
Yet on test scenes, as shown in \cref{fig:visu_recons}, for \ZipNeRFwoRGB{} the inversion model is able to recover the coarse geometry and shading, and sometimes even 
texture and fine objects, with good fidelity. 
This shows that this kind of information remains deeply embedded in the internal representation of NeRFs 
trained with RGB supervision. Hence,  
removing the RGB branch (as done for the \ZipNeRFwoRGB{} baseline) is not sufficient to make it privacy-preserving.
This is further confirmed by the quantitative evaluation provided in \cref{tab:recons_privacy}. 
This lack of privacy shown here for ZipNeRF also applies to any NeRF, including NeRF-based visual localization methods~ \cite{MoreauX23CROSSFIRECameraRelocImplicitRepresentation,chen2023refinement,pietrantoni2024jointnerf,zhou2024nerfect}. Indeed, these NeRF models are optimized with the same photometric loss with RGB ground truth images, share similar underlying rendering equations and have common architectural elements such as MLPs predicting color and geometry. 
In contrast, our 
\PPNeSF{} provides a much higher degree of privacy than \ZipNeRFwoRGB{} both qualitatively and quantitatively.

\section{Privacy-preserving NeSF (\PPNeSF)}
\label{sec:ppnefs}

We next introduce our
privacy-preserving Neural Segmentation Field (\PPNeSF), a neural implicit field encoding 3D segmentation and geometry. 
Rather than training \PPNeSF{} with RGB images, we use segmentations derived from the images as supervision, \ie, no actual images are shown to \PPNeSF{} during training. 
Intuitively, this prevents \PPNeSF{} from having access to potential private information. 
The segmentation algorithm is learned jointly with the field.

\subsection{Architecture}
\label{par:prelim}

Our \PPNeSF{} model,  
shown in \cref{fig:pipeline},
is composed of three modules: a Geometric Field \GF, a Segmentation Field 
\SF{} conditioned on \GF, and an Image Encoder \IE. To enable training from scratch, 
a fourth module, the Feature Field \FF{}, is used solely during training and removed afterwards.

\PARR{The Geometric Field (\GF)} 
is based on the
ZipNeRF \cite{barron2023zip} architecture. 
However, any other \NIF{} could be used as the geometric background model.
Given a ray emanating from a pixel of a camera, a set of 3D points is sampled on the ray, colors are predicted $c_i$, $i\in \{1, ..,n\}$ for each point, and the color of the pixel can be rendered through alpha composition via 
\begin{align}
\hat{C} = \sum_{i=1}^n T_i \alpha_i c_i. 
\end{align}
Here, $T_i$ is the accumulated transmittance at sample $i$ and $\alpha_i$ is the discrete opacity value at the sample 
(see \cite{barron2023zip} for more details). 
To ensure privacy, no photometric supervision is used. 
Hence, 
we remove the RGB branch.
Using alpha composition as above, other representations
such as segmentation logits or features may also be rendered. 
Compared to RGB images, segmentation labels do not provide a strong supervision signal. 
Thus, we add an auxiliary depth loss (see \cite{yu2022monosdf}) to provide further geometric priors.

\begin{figure*}[t!]
\centering
\includegraphics[width=\linewidth]{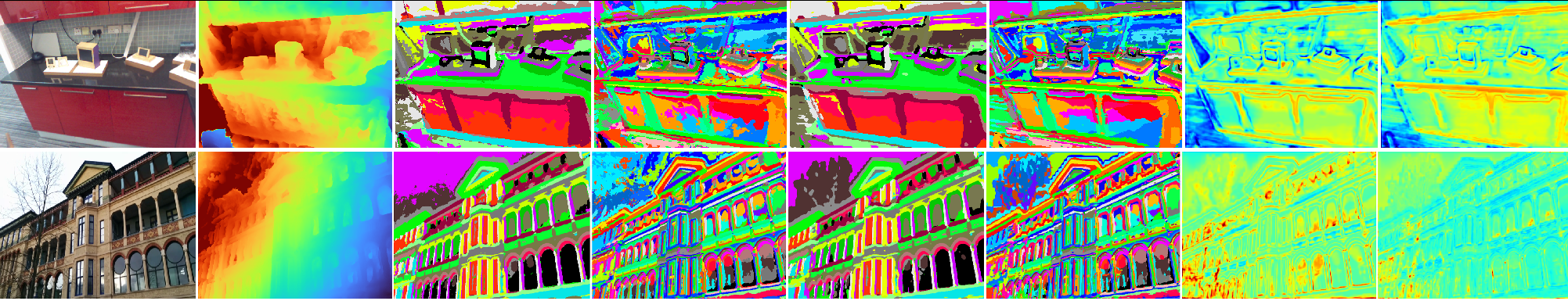}
\caption{From left to right: original image, rendered depth, coarse image-based segmentation $s_c^{2D}$, fine image-based segmentation $s_f^{2D}$, coarse rendered segmentation $s_c^{3D}$, fine rendered segmentation $s_f^{3D}$, coarse uncertainty map $u^{2D}_c$ and fine uncertainty map $u^{2D}_f$.
Rendered and image-based segmentation are well aligned which allows for precise visual localization.}
\label{fig:visu_segs}
\end{figure*}

\PAR{The Segmentation Field (\SF)}
We introduce a segmentation head that maps each 3D point to a segmentation logit. Segmentation logits along a ray may be rendered by alpha composition, yielding the segmentation map $s^{3D}$.
This head is conditioned on the first MLP of the geometry field, allowing geometry optimization through segmentation supervision. 

\PARR{The Image  Encoder (\IE)}
is composed of a transformer feature backbone and a decoder. 
The decoder outputs a
segmentation map $s^{2D}$ and an uncertainty map $u^{2D}$ from multi-resolution features from the feature backbone.

\PARR{The Feature Field (\FF)}
is only used to help train  
the Segmentation Field \SF{}. 
It is deleted after training 
to maintain  
privacy. 
The 3D feature field is used to render volumetric features (using the same anti-aliasing techniques as ZipNeRF) 
by alpha composition using the opacity weights from \GF{}.
Gradients are detached from these weights to avoid privacy information leaking into the Segmentation Field (\SF).

\subsection{Self-supervised training}

To learn aligned segmentations between the Image Encoder \IE{} and the Segmentation Field \SF{} along with the Geometry Field \GF, we first define the segmentation classes by clustering features in a learned embedding space shared between the 
Image Encoder and the Feature Field \FF. 

Below, we describe how the shared feature embedding space is learned (\cref{sec:feat_emb}); 
how we obtain two sets of prototypes  (cluster centres) in this space; and how \PPNeSF{} is trained 
(\cref{sec:opt})  
by alternatively deriving segmentation label targets from the feature embeddings and using these targets in the 
 learning objective to jointly optimize \SF{}, \IE{}, and \GF.

\subsubsection{Learning the feature embedding space}
\label{sec:feat_emb}

\PAR{Feature alignment} We first learn a joint embedding space between the Encoder \IE{} and 
features rendered from \FF.
Given an image $I$ with pose $P$, the Encoder extracts a per-pixel feature map $F^{2D} \in \Re^{D \times H \times W}$. 
Given the pose $P$, we render a per-pixel  
volumetric feature map $F^{3D} \in \Re^{D \times H \times W}$ from the Feature Field \FF~using alpha composition.
Our aim is to ensure that the 2D/3D features maps are pixel-aligned, while encouraging feature discriminativeness through the following contrastive loss~\cite{vandenOordX18ReprLearWithContrastivePredCoding}: 
\begin{align}
L_{NCE} = -\frac{1}{2N}  \sum_{i=1}^N  \log \left( \frac{\exp{\left(F^{3D}_{u_i} F^{2D}_{u_i}/\tau \right)}^2}{A} \right) \enspace.
\label{eq:Lnce}
\end{align}
Here, $\{u_i\}_{i=1}^N$ are  randomly sampled pixels, 
$\tau$ is a temperature parameter, 
and A is a normalizing factor (see supp.~mat.). 
In order to avoid the contamination of \PPNeSF{} 
with texture and color information, and hence compromising privacy, we do not backpropagate the gradients between \FF{} and 
(\SF,\GF).
Embedding 2D/3D features in the same embedding space leverages the scale-awareness and viewpoint-invariance of the 3D Feature Field \FF, while benefiting from the inductive biases of the transformer-based 2D Encoder \IE. 
As segmentations targets are derived from the embedding space, they also inherit these properties.

\PAR{Prototypes} In the embedding space, we maintain two \textit{aligned} sets\footnote{The need for two separate sets of prototypes arises because our model is trained from scratch and initialized randomly. At the beginning of the training process, the volumetric and image-based features are not yet aligned. Hence, using a shared set of prototypes that jointly partition both feature spaces makes less sense and the training would be ineffective.}
of $K$ prototypes, one for the image-based features $\{p^{2D}_{k}\}_{k=1}^{K}$ and one for the rendered features $\{p^{3D}_{k}\}_{k=1}^{K}$, with $K$ being the number of segmentation classes. 
They implicitly define the latent classes in the feature embedding space and will be used to derive segmentation targets by mapping features to prototypes based on their similarities in the corresponding feature space.
Since the prototypes capture the whole scene information, mapping features from a single image to the prototypes 
will ensure view-consistent and discriminative segmentation targets. Hence they provide a good learning signal to optimize both the Segmentation Field \SF{} and the Image Encoder \IE.

\subsubsection{Optimizing \PPNeSF{}}
\label{sec:opt}
To ensure privacy, \PPNeSF{} is solely optimized through a cross-entropy loss. 
Each training iteration consists of two alternating steps: Computing the target segmentation label distribution Q and optimizing the loss with these labels.

\PAR{Objective function}
Given a target segmentation label distribution Q, we aim to minimize the following cross entropy loss with regard to the \PPNeSF{} parameters (\GF{}, \SF{}, and  \IE): 
\begin{align}
 \min \sum_{i=1}^N \sum_{k=1}^{K} q(k|u_i) \log{\left(s^{2D}(k|u_i)\cdot s^{3D}(k|u_i)\right)} \enspace . 
\label{eq:ce}
\end{align}
Here, $\{u_i\}_{i=1}^N$ is a set of uniformly sampled pixels, and 
$s(.|u_i)$ is the softmax prediction score over the $K$ segmentation classes of the image encoder \IE{} ($s^{2D}$)  and of the rendered segmentation field \SF{} ($s^{3D}$), respectively.
We effectively enforce that both the image-based segmentations and rendered segmentations align with the same target distribution $Q$ so that segmentation-based pose alignment may be run after training. Note that the geometry of the scene is also optimized through this objective function.

\begin{table*}[tt]
\scriptsize
\resizebox{\linewidth}{!}{
  \begin{tabular}{cl|c||c|c|c|c|c|c|c}
   &  Model & Privacy & Chess & Fire & Heads & Office & Pumpkin & Redkitchen & Stairs  \\
   \hline
 \multirow{2}*{\begin{sideways} SBM \end{sideways}}  
 & HLoc~\cite{SarlinCVPR19FromCoarsetoFineHierarchicalLocalization} & x & 0.8/0.11/100 & 0.9/0.24/99 &  0.6/0.25/100 & 1.2/0.20/100 & 1.4/0.15/100 & 1.1/0.14/99  &2.9/0.80/72 \\
  & DSAC*\cite{brachmann2021visual} & ? & 0.5/0.17/100 &  0.8/0.28/99 & 0.5/0.34/100 & 1.2/0.34/98 & 1.2/0.28/99 & 0.7/0.21/97 & 2.7/0.78/92 \\
  \hline
  \multirow{7}*{\begin{sideways} RBFM  \end{sideways}}
   & NeFeS~\cite{chen2023refinement} (DFNet \cite{ShuaiECCV22DFNetEnhanceAPRDirectFeatureMatching}) & x & 2/0.79 & 2/0.78 & 2/1.36 & 2/0.60 & 2/0.63 & 2/0.62 & 5/1.31\\
&     SSL-Nif (DV)~\cite{pietrantoni2024jointnerf} & x & 1/0.22/93 & 0.8/0.28/91 & 0.8/0.49/71 & 1.7/0.41/81 & 1.5/0.34/86 & 2.1/0.41/75 & 6.5/0.63/49 \\
 & NeRFMatch (DV)~\cite{zhou2024nerfect} & x & 0.9/0.3 & 1.3/0.4 & 1.6/1.0 & 3.3/0.7 & 3.2/0.6 & 1.3/0.3 & 7.2/1.3 \\
& MCLoc \cite{trivigno2024unreasonable} &  x & 2/0.8 & 3/1.4  & 3/1.3 & 4/1.3  & 5/1.6 & 6/1.6 & 6/2.0  \\
& GSplatLoc \cite{sidorov2024gsplatloc} (GSplatLoc)& x &  0.43/0.16 & 1.03/0.32 & 1.06/0.62 & 1.85/0.4 & 1.8/0.35 & 2.71/0.55 & 8.83/2.34  \\
& GS-CPR \cite{liu2024gsloc} (DFNet)& x & 0.7/0.20 & 0.9/0.32 & 0.6/0.36 & 1.2/0.32 & 1.3/0.31 & 0.9/0.25 & 2.2/0.61 \\
& STDLoc \cite{huang2025sparse} & x & 0.46/0.15 & 0.57/0.24 & 0.45/0.26 & 0.86/0.24 & 0.93/0.21 & 0.63/0.19 & 1.42/0.41 \\
\hline
\multirow{2}*{\begin{sideways} PP \end{sideways}} 
 & GSFF \cite{pietrantoni2025gaussian} (DV) & \checkmark & \underline{0.8/0.28}/\textbf{96} & \underline{0.8/0.33/94} & \underline{1.0/0.67}/\textbf{90} & \underline{1.5/0.51/90} & \underline{2.0/0.50/78} & \underline{1.2/0.33/85} & \textbf{28.2}/\underline{0.98}/\textbf{29} \\
  & \textbf{\PPNeSF{}} (DV) & \checkmark & \textbf{0.6/0.17}/\underline{95} & \textbf{0.7/0.28/98} & \textbf{0.6/0.36}/\underline{85} & \textbf{1.3/0.35/98} & \textbf{1.9/0.48/89} & \textbf{0.9/0.24/92} & \underline{29}/\textbf{0.7}/\underline{27}  \\
     \hline
     \bottomrule
    \end{tabular}
}
\caption{Localization results for the 7Scenes dataset (SfM-based GT from \cite{BrachmannICCV21OnTheLimitsPseudoGTVisReLoc}), comparison against rendering-based feature methods (RBFM). Median pose error (cm.) ($\downarrow$)/ Median angle error (°) ($\downarrow$)/ Recall at 5cm/5° ($\%$) ($\uparrow)$. Our main focus is comparing with other privacy-preserving (PP) approaches and we highlight the \textbf{best} and \underline{second best} approaches in this category.}
      \label{tab:7scenesWithpGT}
    \end{table*}

\PAR{Deriving target label distributions $Q$}
\label{par:labelling}
At each iteration, given a training image, we first derive 
a segmentation label distributions $Q$ as follows.
Using Frobenius inner product notations,  \cref{eq:ce} can be rewritten as $\min \langle Q, -log S \rangle$, where $Q_{ik} = q(k|u_i)$ and $S_{ki} = s^{2D}(k|u_i)) s^{3D}(k|u_i)$. The target distribution $Q$ is the minimizer of this equation.
By adding an entropy regularization term on $Q$ and relaxing $Q$ to belong to the transportation polytope~\cite{distances2013lightspeed}, the former equation can be rewritten as an optimal transport (OT) problem. 
Thus, we seek a feature-to-class mapping $Q$ that maximizes an objective of the form 
\begin{align}
    \max_{Q \in U(\frac{1}{N},\frac{1}{K})} \text{Trace}(Q (-log S)^t) + \lambda h(Q) \enspace .
    \label{eq:ot}
\end{align}
 $Q$ can be efficiently computed with the iterative Sinkhorn-Knopp algorithm~\cite{distances2013lightspeed}. The resulting $Q \in \Re^{N \times K}$ is used to optimize \cref{eq:ce} with regard to \GF, \SF{}, and \IE.

To derive $Q$, we rely on the softmax scores\footnote{It can be seen as Nearest Class  Mean \cite{MensinkPAMI13DistanceBasedGeneralizingToNewClasses}, where each prototype represents the corresponding class mean.}
obtained from feature-prototypes similarities. Thus $S \in \Re^{K\times N}$ becomes  
\begin{align*}
S_{ki}  = 
\exp{\left( (F^{2D}_{u_i} p_k^{2D} + F^{3D}_{u_i} p_k^{3D})/\tau \right) }/B \,
\end{align*}
with $B$ being a normalization factor.
Despite using only $N$ sampled pixels per iteration, we compare the features $F_u$ with prototypes $p_k$, which encapsulate the entire scene’s content. This approach ensures that the resulting labels are discriminative, diverse, and viewpoint-consistent, which are key properties needed for optimizing the implicit field.

Finally, we obtain pixel-wise assignments as $v=\argmax_k(Q)$. For each class $k$ we can then define the set of assigned pixels as $A(k) = \{ i | v(i)=k\}$. 
The prototypes are then updated through an Exponential Moving Average (EMA)
scheme \cite{BrownB63SmoothingForecastingandPredDTS} based on each pixel's assignment 
$$\textstyle{
p^{2D}_{k} = \mu p^{2D}_{k} + (1-\mu) ( \beta \frac{1}{|A(k)|} \sum_{i\in A(k)} F^{2D}_i }$$
$$\textstyle{+ (1 - \beta) \frac{1}{|A(k)|} \sum_{i\in A(k)} F^{3D}_i) } \enspace , $$ 
with $\beta$  being scheduled from 0 to 0.5 over the course of the training (3D prototypes are updated similarly).

\subsection{Regularization}
Replacing photometric supervision with segmentation supervision makes the optimization less stable as segmentation contain less information about the scene and no texture. To facilitate the training of \PPNeSF{}, we introduce a coarse-to-fine hierarchical segmentation scheme and we model
segmentation uncertainty. 
More details are provided in the  \cref{sec:arch}.

\PAR{Hierarchical segmentation}
\label{par:hierar}
To capture different granularities and to provide finer complementary supervision information, we introduce a fine level of segmentation. 
Concretely, each of the previous clusters is divided into $n$ sub-clusters yielding a set of $K_{f}=n\cdot K$ fine prototypes for the 2D and 3D modalities each.
A second segmentation head of dimension $K_f$ is added to both the implicit model \SF{} and the Encoder \IE.
The fine prototypes are updated by EMA similar to the update of the coarse prototypes.
To further promote intra-class compactness and separability within the embedding space, inspired by \cite{zhang2022use}, we 
use a hierarchical contrastive loss between pixels and prototypes, which also enforces the hierarchy defined above.

\PAR{Uncertainty}
\label{par:unc}
Our self-supervised segmentation targets are affected by uncertainty due to the labelling procedure. 
As they are used as the main supervision signal, this can destabilize the learning process.
Therefore, following \cite{kendall2017uncertainties}, we give the encoder the possibility to predict heteroscedastic uncertainty and attenuate the cross entropy loss for uncertain samples. 
These uncertainties are further used during localization to down-weight ambiguous pixels.

\section{Experiments}
\label{sec:exps}
This section first presents our full visual localization pipeline as well as details regarding the evaluation followed by comprehensive visual localization 
experiments and ablation studies. \cref{fig:visu_segs} shows resulting segmentations after training. Comprehensive details regarding training and evaluation may be found in the Appendix.

\PAR{Visual localization pipeline}
\label{par:loc_pip}
Given a query image with unknown pose, 
following standard practice, we first estimate an initial pose $P_0$ 
via image retrieval as the pose of the closest database image based on global descriptor similarities, \eg, using DenseVLAD (DV) \cite{ToriiPAMI18247PlaceRecognitionViewSynthesis} or SegLoc (Sgl) \cite{pietrantoni2023segloc} depending on the dataset. 
Starting from $P_0$, and given a randomly sampled set of query image pixels, segmentation labels $s^{3D}$ are rendered from \PPNeSF{} while 2D segmentation labels $s^{2D}$ are extracted from the query image through the encoder.
The cross entropy is then minimized with regard to the pose as
$$\textstyle{ \min_{P \in SE(3)} \sum_{i=1}^N \sum_{k=1}^{K}  s_m^{2D}(k|u_i)log(s_m^{3D}(k|u_i)) \enspace .}$$ 
Here, $m \in\{c,f\}$ is the segmentation level (coarse or fine). Segmentation logits are weighted with the sampled uncertainty.
By backpropagating through the implicit model, the pose is updated and this optimization process is iteratively repeated for a fixed number of iterations.
We sequentially repeat this process for the coarse and the fine segmentation as to increase the convergence basin of the pose estimation. 

In a cloud-based localization setting, the client computes the 
global descriptor and segmentation for the query image and sends both to the server. 
The server stores global descriptors and poses for the database images, as well as the \PPNeSF{} model. 
Since training a \PPNeSF{} involves RGB images (even if the \PPNeSF{} model never sees RGB images as input), the training needs to be done locally by a client. 
Models are independently trained per scene. We set the number of coarse classes to $K=20$ and choose to $n=5$ for a total number of $K_f=100$ fine classes. 
We found that the choices provide a good compromise between performance and training/inference speed (see \cref{tab:trainingtime}).

\begin{table}[t!]
  \centering
  \resizebox{\linewidth}{!}{
  \begin{tabular}{cl|c||c|c|c|c}
   &  Model& Privacy & King's & Old & Shop & St. Mary's \\
    \hline
     \multirow{3}*{\begin{sideways}  SBM \end{sideways}}
    & HLoc~\cite{SarlinCVPR19FromCoarsetoFineHierarchicalLocalization} & x & 0.11/0.20 & 0.15/0.31 & 0.04/0.20 & 0.07/0.24  \\
 &    PixLoc~\cite{SarlinCVPR21BackToTheFeature}  (NV)   & x & 0.14/0.24 & 0.16/0.32 & 0.05/0.23 & 0.10/0.34  \\
 &   DSAC*\cite{brachmann2021visual} & ? & 0.15/0.30& 0.21/0.40 & 0.05/0.30 & 0.13/0.40 \\
   \hline
     \multirow{10}*{\begin{sideways} RBFM \end{sideways}}
  &  FQN~\cite{GermainCWPRWS21FeatureQueryNetworks} (FQN-PnP) 
 & x & 0.28/0.40 & 0.54/0.80 & 0.13/0.60 & 0.58/2.00  \\
 &   CROSSFIRE~\cite{MoreauX23CROSSFIRECameraRelocImplicitRepresentation} (DV)   & x & 0.47/0.70 & 0.43/0.70 & 0.20/1.20 & 0.39/1.40 \\
   & NeFeS~\cite{chen2023refinement} (DFNet) & x & 0.37/0.62 & 0.55/0.90 & 0.14/0.47 & 0.32/0.99   \\
    &  SSL-Nif (DV) \cite{pietrantoni2024jointnerf} & x & 0.32/0.40 & 0.3/0.49 & 0.08/0.36 & 0.24/0.66  \\
     & NeRFMatch (DV)~\cite{zhou2024nerfect} & x & 0.13/0.20 & 0.19/0.40 & 0.08/0.40 & 0.07/0.30 \\
     & MCLoc \cite{trivigno2024unreasonable} & x & 0.31/0.42 & 0.39/0.73 & 0.12/0.45 & 0.26/0.88  \\
     &  GS-CPR \cite{liu2024gsloc} (DFNet) & x & 0.26/0.34& 0.48/0.72 & 10/0.36 & 0.27/0.62 \\
      & GSplaLoc \cite{sidorov2024gsplatloc} & x & 0.27/0.46 & 0.20/0.71 & 0.05/0.36 & 0.16/0.61 \\
      & STDLoc \cite{huang2025sparse} & x & 0.15/0.17 & 0.12/0.21 & 0.03/0.13 & 0.05/0.14 \\ 
      \hline
       \multirow{5}*{\begin{sideways} PP \end{sideways}} 
   & GoMatch \cite{zhou2022geometry} & \checkmark & 0.25/0.64 & 2.83/8.14 & 0.48/4.77 & 3.35/9.94 \\
   &  DGC-GNN \cite{wang2023dgc} & \checkmark & \textbf{0.18}/0.47 & 0.75/2.83 & 0.15/1.57 & 1.06/4.03  \\
   & SegLoc \cite{pietrantoni2023segloc} & \checkmark & 0.24/\textbf{0.26} & 0.36/0.52 & 0.11/0.34 & 0.17/\underline{0.46} \\ 
    & GSFF \cite{pietrantoni2025gaussian} (DV)& \checkmark & 0.24/0.39 & \underline{0.26/0.49} & \textbf{0.05/0.27} & \underline{0.13}/0.48 \\
   &  \textbf{\PPNeSF} (DV) & \checkmark &\underline{0.20/0.30} & \textbf{0.19/0.39} & \underline{0.08/0.39} & \textbf{0.09/0.35} \\
     \hline
 \hline
    \bottomrule
    \end{tabular}
    }
    \caption{Localization results on  Cambridge Landmarks, comparison against rendering-based feature methods (RBFM) and privacy-preserving methods (PP). Median pose error (m.) ($\downarrow$), Median angle error (°) ($\downarrow$). \textbf{Bold} best numbers, \underline{underline} second best numbers among privacy-preserving methods.}
      \label{tab:cambridge}
    \end{table}

\PAR{Evaluation protocol} We evaluate our \PPNeSF{} localization pipeline on 
three real-world datasets:
7Scenes~\cite{ShottonCVPR13SceneCoordinateRegression}, Indoor6~\cite{do2022learning}, and Cambridge Landmarks~\cite{KendallICCV15PoseNetCameraRelocalization}.
To evaluate localization accuracy we use the standard 
median translation error (in m or cm), median rotation angle error (°), and the recall at 5cm/5°
in the case of indoor scenes.

\begin{table*}[tt!]
\scriptsize
\resizebox{\linewidth}{!}{
  \begin{tabular}{lr|c|c|c|c|c|c}
      & PP & scene1 & scene2a & scene3 & scene4a & scene5 & scene6  \\
    \hline
    HLoc \cite{SarlinCVPR19FromCoarsetoFineHierarchicalLocalization} & x & 3.2/0.47/64.8 & 3.9/0.76/60.6 & 2.1/0.37/81.0 & 3.3/0.47/70.6 & 6.1/0.86/42.7 & 2.1/0.42/79.9  \\
    DSAC* \cite{brachmann2021visual} & ?& 12.3/2.06/18.7 & 7.9/0.9/28.0 & 13.1/2.34/19.7 & 3.7/0.95/60.8 & 40.7/6.72/10.6 & 6.0/1.40/44.3  \\
    \hline
    SegLoc (Sgl) \cite{pietrantoni2023segloc} & \checkmark &\textbf{3.9}/0.72/51.0 &  \textbf{3.2/0.37}/\underline{56.4} & 4.2/0.86/41.8 & \underline{6.6}/1.27/33.84 & \underline{5.1/0.81/43.1} & \underline{3.5}/0.78/34.5 \\
    GSFF (Sgl) \cite{pietrantoni2025gaussian} & \checkmark &  \underline{4.6/0.67/53} & 4.1/0.33/49 & \underline{4.1/0.74/55} & \textbf{4.0/0.61}/\underline{58} & 6.6/0.92/38 & 3.7/\underline{0.75/57} \\
    \PPNeSF{} (Sgl) & \checkmark & \textbf{3.9/0.63/61.5} & \underline{3.7/0.38}/\textbf{70.0} & \textbf{3.2/0.67/67.0} & \textbf{4.0}/\underline{0.97}/\textbf{62.0} & \textbf{4.7/0.72/54.2} & \textbf{3.3/0.56/62.2} \\
    \hline
    \bottomrule
    \end{tabular}}
\caption{Localization results on the Indoor6 dataset, comparison with privacy-preserving (PP) baselines. Median pose error (cm.) ($\downarrow$)/ Median angle error (°) ($\downarrow$)/ Recall at 5cm/5° ($\%$) ($\uparrow)$. \textbf{Bold} best numbers, \underline{underline} second best numbers among PP methods.}
      \label{tab:indoor6}
    \end{table*}

\subsection{Evaluation results}
We primarily compare our method against other privacy-preserving (PP) visual localization methods. 
For the sake of completeness, we also compare against non-privacy-preserving rendering-based feature methods (RBFM). Structure-based methods (SBM), SfM ~\cite{SarlinCVPR19FromCoarsetoFineHierarchicalLocalization,SarlinCVPR21BackToTheFeature} or regression\footnote{Note that no study have been conducted in the literature on how privacy preserving are the 3D regression models.} \cite{brachmann2021visual}, despite not focusing on privacy are added for reference.  

We report results for 7Scenes with SfM ground truth~\cite{BrachmannICCV21OnTheLimitsPseudoGTVisReLoc} (baselines reporting only results with DSLAM ground truth are not added) in \cref{tab:7scenesWithpGT}, for Cambridge Landmarks in \cref{tab:cambridge}, and for Indoor6 in \cref{tab:indoor6}.

\PAR{Comparisons with privacy-preserving methods (PP)}
We compare \PPNeSF{} to privacy-preserving approaches. In particular, to descriptor-free approaches, GoMatch~\cite{zhou2022geometry} and DGC-GNN~\cite{wang2023dgc}, and methods that use segmentation labels to preserve privacy (SegLoc~\cite{pietrantoni2023segloc} and GSFF~\cite{pietrantoni2025gaussian}).  
We can see from Tables.~\ref{tab:7scenesWithpGT}-\ref{tab:indoor6} that
\PPNeSF{} achieves significantly better performance than descriptor-free approaches. Also, 
it outperforms SegLoc~\cite{pietrantoni2023segloc} and on average outperforms prior state-of-the-art GSFF~\cite{pietrantoni2025gaussian}, showing consistent gains
over all the evaluated datasets.

\PAR{Comparisons with rendering-based feature methods (RBFM)}
RBFM~\cite{zhou2024nerfect,pietrantoni2024jointnerf,chen2023refinement,MoreauX23CROSSFIRECameraRelocImplicitRepresentation,liu2024gsloc,trivigno2024unreasonable,sidorov2024gsplatloc,huang2025sparse} use the rendered features 
for matching or alignment to perform pose refinement.
Hence, these methods are not privacy-preserving as these features contain fine-grained details. 
Further, the NeRF models of ~\cite{zhou2024nerfect,pietrantoni2024jointnerf,chen2023refinement,MoreauX23CROSSFIRECameraRelocImplicitRepresentation} are not
privacy preserving as they are optimized with a photometric loss and by design allow image rendering.
Despite providing a higher degree of privacy and often starting from  a poorer initial pose, 
\eg, DenseVlad (DV) \vs more accurate initialization such as DFNet/GSplatLoc, our \PPNeSF{} outperforms most of the rendering-based feature methods in terms of accuracy. On 7Scenes (see \cref{tab:7scenesWithpGT}), \PPNeSF{} remains competitive with the SOTA Gaussian Splatting-based STDLoc~\cite{huang2025sparse} on most of the scenes 
(except \emph{Stairs}, where our  segmentation-based approach struggles compared to methods using more informative deep features due to the texture-less and repetitive nature of the scene).
\PPNeSF{} obtains also very accurate results on Cambridge Landmarks (see \cref{tab:cambridge}), outperforming 
all but the 
non-privacy-preserving
STDLoc~\cite{huang2025sparse} and NeRFMatch~\cite{zhou2024nerfect} methods.

\subsection{Ablation studies}
\label{sec:ablations}

\cref{tab:abl_camb} ablates different components of our model on the Cambridge Landmarks dataset. 
The top two rows show the advantage of the hierarchical segmentation, which increases discriminativeness and the convergence basin.
Assigning labels based on L2 distances between features and prototypes (\textit{L2 attribution}) for 2D and 3D separately leads to the highest drop in performance, emphasizing the importance of the joint 2D/3D OT-based labelling process.
Using two sets of prototypes instead of one 
leads to
faster convergence and higher accuracy 
in general (
(\textit{Shared Prototypes}). 
Obtaining them by directly clustering the feature embedding space (\textit{K-means}) instead of the EMA updating scheme, leads to a further drop in accuracy. 
Training without  
the 
hierarchical loss (\textit{No $L_{hierar}$}) induces a drop in accuracy, which suggests that the loss helps having a more separated feature embedding space which leads to more consistent segmentations.
The uncertainty modeling (\textit{No Uncertainty})
improves performance by filtering out ambiguous pixels.

\begin{table}[t!]
\scriptsize
  \centering
  \resizebox{\linewidth}{!}{
  \begin{tabular}{l|c|c|c|c|c}
   Model & King's & Old & Shop & St. Mary's & Average\\
    \hline
       Coarse only& 28.6/0.47 & 29.6/0.55 & 9.8/0.41 & 15.3/0.71 & 20.9/0.54\\
       Fine only & 25.4/0.41 & 33.2/0.58 & 8.7/0.38 & 13.9/0.62 & 20.3/0.50\\
       L2 attribution & 23.1/0.32 & 65.2/0.90 & 11.5/0.59 & 17.7/0.64 & 29.4/0.61\\
       Shared prototypes & 22.7/0.34 & 23.2/0.37 & 8.2/0.38 & 10.7/0.36 & 16.2/0.36\\
       K-means & 26.6/0.44 & 29.9/0.56 & 11.2/0.49 & 14.6/0.52 & 20.6/0.50\\
       No $L_{hierar}$ & 21.0/0.31 & 21.7/0.40 & 8.4/0.39 & 9.9/0.36 & 15.3/0.37\\
       No Uncertainty & 21.2/0.31 & 20.9/0.40 & 8.5/0.41 & 11.1/0.38 & 15.4/0.38\\
       \hline
       \PPNeSF{} & 20.6/0.30 & 19.9/0.39 & 8.2/0.39 & 9.6/0.35 & 14.6/0.36 \\
    \hline
    \bottomrule
    \end{tabular}
    }
    \caption{Ablations of different components of our pipeline on Cambridge Landmark. Median position error (cm.) ($\downarrow$) / median rotation error (°) ($\downarrow$). NetVlad (NV) initialization.}
    \label{tab:abl_camb}
    \end{table}

\section{Conclusion} 
This paper investigates the privacy preservation of Neural Radiance Fields (NeRFs) in the context of visual localization.
We propose a novel NeRF inversion privacy attack and a corresponding privacy evaluation protocol. It shows 
that RGB supervision commonly used to optimize the NeRFs results in the encoding of fine-grained information within the geometric part of the model. 
We thus propose an alternative privacy-preserving neural implicit field, where we replace the RGB supervision by hierarchical segmentation label supervision. 
To mitigate the potential loss in the quality of the underlying neural field caused by removing RGB supervision, we carefully design a method to learn robust and discriminative segmentations along with associated uncertainties in an unsupervised manner.

We demonstrate on several datasets that our resulting privacy segmentation-based approach \PPNeSF{} not only outperforms other privacy-preserving methods but remains competitive even with non privacy-preserving alternatives.

{
   
    \small
    \bibliographystyle{ieeenat_fullname}

}

\vspace{1cm}
 {\Large \bf APPENDIX} \\
 
 \appendix

This section is structured as follows. 
In \cref{sec:arch}
we first provide  
comprehensive details regarding the \PPNeSF{}  architecture, 
including a detailed description of 
the image-based encoder \IE, the feature field \FF, and the underlying neural implicit field. In addition we give  more details regarding the hierarchical scheme 
and the uncertainty computation. 
In \cref{sec:training},  first we detail how the  model is trained, 
then, we provide  further details on the pose refinement, the relation between training/runtime cost vs. performance, and study the impact of the number of segmentation classes considered on the localization accuracy.
Finally, in \cref{sec:more_privacy_exps} we expand upon the privacy aspect.
First, 
we  describe the inversion  model used for the privacy attack and describe our training and evaluation protocol for the privacy attack. 
 Then, in \cref{sec:further_baselines} we provide additional privacy baseline experiments and 
 in \cref{sec:segmprivacy}
 an analysis on why using segmentations
 allows high level privacy preservation. 

\section{Details on the \PPNeSF{}  architecture}
\label{sec:arch}
\PARR{Image-based encoder \IE}
is composed of a feature backbone and a decoder.
The backbone is a SWIN-t transformer \cite{liu2021swin} (patch size: 2, embed dim: 96, depths: $ \{2, 2, 6, 2\} $,  num heads = $\{3, 6, 12, 24\}$) which outputs a set of 4 feature maps with strides (2, 4, 8, 16). 
The feature map with stride 2 has a dimension of 96. 
It is processed through a feature head consisting of three conv2D layers (internal dimension 128) with ReLU activation and one upsampling layer such that the feature map is pixel aligned. 
This resulting feature map has a dimension of 96 and is used for the joint embedding space learning. 
The multi-scale feature maps are processed with a convolutional feature pyramid type of network \cite{lin2017feature} to output two final feature maps of stride 4 and 2. 
They are processed through segmentation heads (three conv2D layers with ReLU activation and internal dimension of 128, one upsampling layer of scale 4 or 2) which yield the pixel aligned coarse and fine segmentation maps with respectively $K$ and $K_f$ classes. 

These final feature maps are processed through uncertainty heads (three conv2D layers with ReLU activation and internal dimension of 128, one upsampling layer of scale 4 or 2) to obtain the pixel aligned coarse and fine uncertainty maps with respectively $K$ and $K_f$ dimensions. 
The gradients are detached from the input of the feature head, coarse and fine uncertainty heads as to not destabilize the segmentation representation learning. 
No pre-trained weights are used to initialize the encoder.

\PARR{Feature field \FF} 
consists of hash tables with 10 levels with resolution ranging from 16 to 16,384. 
The feature dimension per level is set to 4 with a log hashmap size of 21. 
Following \cite{barron2023zip}, a conical section of a ray consists of 6 points and is encoded through the hash tables. 
The encoded features are weighted and averaged. 
This encoding makes the 3D feature scale-aware which is beneficial for scenes with large viewpoint changes. 
The resulting feature is then concatenated with the scale feature (see \cite{barron2023zip} for more details) and processed by a MLP consisting of 3 linear layers (internal dimension 128) with ReLU activations. 
The output feature has a dimension of 96 to match with the encoder-based feature dimension. 
Feature rendering uses the opacity weight from the main neural implicit field. 
Gradients are detached from these weights.

\PARR{\NIF} (neural implicit field) is a representation of a 3D scene whose underlying geometry is encoded through MLPs. 
Additional information such as radiance, semantics, or features may also be stored and encoded. 
NeRFs~\cite{mildenhall2020nerf} are a special form of neural implicit fields which store radiance and use volumetric rendering~\cite{Kajiya1984VolumeRendering} to optimize the neural fields solely using RGB images with known intrinsics and camera poses. 
Follow-up work tackle training efficiency by introducing 3D structures such as hash tables, octrees, or grids \cite{M_ller_2022,chen2022tensorf,yu2021plenoctrees, wang2022fourier, yu2021plenoxels,chen2023factor}, reducing the number of required training views by adding additional regularization or geometric priors \cite{yu2022monosdf,niemeyer2022regnerf,truong2023sparf,wang2023sparsenerf}, and improving the quality by reducing aliasing \cite{barron2021mipnerf,barron2023zip,nam2024mip}. 

\PARR{\PPNeSF} uses the ZipNerf architecture as the background neural implicit field.\footnote{Our implementation is based on \url{https://github.com/SuLvXiangXin/zipnerf-pytorch.git}.}
Concretely, given an image with known pose, each emitted ray is modeled as a 
conical frustum whose sections can be decomposed into a set of 6 isotropic Gaussians such that they approximate the shape of the conical frustum. 
For a conical section, 
for each level $V_l$ of the grid, a feature is obtained by trilinear interpolation applied at the Gaussian's mean location $x_j$. 
The six resulting features are reweighted $w_{j,l}$ based on how each Gaussian fits into the grid cell and the averaged $$\textstyle{f_l = \mean_j(w_{j,l} trilerp(x_j;V_l) \enspace ,}$$ 
$trilerp$ being the trilinear interpolation operation.
The resulting level features are then concatenated and fed into a shallow MLP to obtain the density and a geometric feature $d_i,g_i = h_{geo}(cat[f_l]_l)$. 
The geometric feature is fed along with the encoded point position to a shallow MLP to obtain the 3D segmentation $$\textstyle{s_i^{3D} = h_{seg}(cat[g,enc(pos)]) \enspace .}$$

This combination of multi-sampling and weighting effectively reduces spatial-aliasing and handles scale variations. 
Segmentations can finally be rendered through alpha composition $s^{3D} = \sum_{i=1}^n T_i \alpha_i s_i^{3D}$,  where $T_i$ is the accumulated transmittance at sample $i$ and $\alpha$ the discrete opacity value at sample $i$. 
Furthermore, Z-aliasing induced by the proposal sampling is handled by deriving an inter-level loss which is smooth with regard to translation along the ray. 
For more details please refer to \cite{barron2023zip}. 

\PARR{PPNeSF's geometric field \GF{}} has similar architecture as in \cite{barron2023zip}, except for the segmentation part.
Two levels of proposal networks are used, their hash tables' maximum resolutions are 512 and 2048. 
The main network uses a hash grid with 10 levels with resolution ranging from 16 to 16,384. 
The feature dimension per level for all hash tables is set to 4 with a log hashmap size of 21. 
The geometric MLP contains two linear layers and ReLU activations (internal dimension 64), scale featurization is used. 
Coarse and fine semantic heads consist of a three layer MLP with ReLU activations (internal dimension 128). 

The number of samples per proposal network is set to 48 while the number of samples for the main network is set to 24. 
Sampling is handled by the proposal networks (see \cite{barron2023zip} for more details).

\begin{figure*}[tt!]
\centering
\includegraphics[width=1.\linewidth]{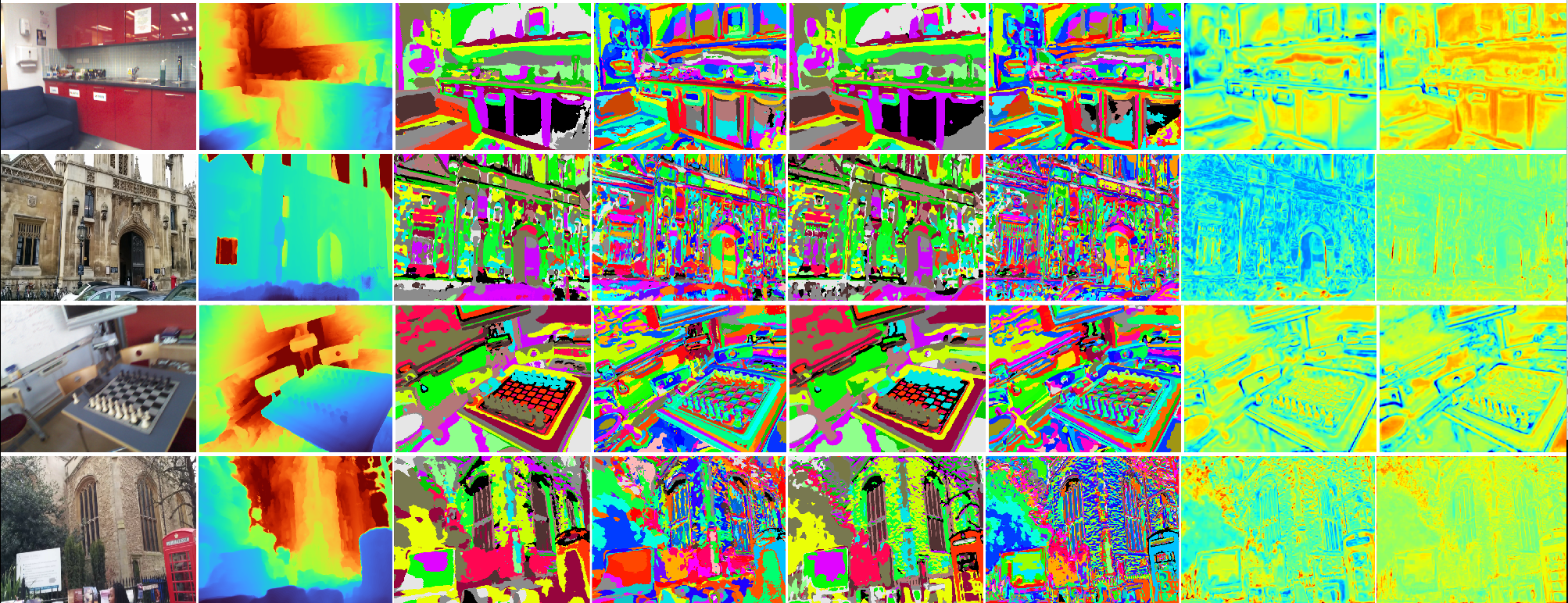}
\caption{From left to right: original image, rendered depth, coarse image-based segmentation $s_c^{2D}$, fine image-based segmentation $s_f^{2D}$, coarse rendered segmentation $s_c^{3D}$, fine rendered segmentation $s_f^{3D}$, coarse uncertainty map $u^{2D}_c$ and fine uncertainty map $u^{2D}_f$.}
\label{fig:more_seg_visu}
\end{figure*}

\PARR{The hierarchical framework} we introduced 
in \cref{sec:ppnefs}, 
provide a finer supervision to \PPNeSF{} and increase the convergence basin during visual localization. 
Here, we describe  how the fine segmentation targets are derived. For each coarse class $k \in [1,K]$, we want to establish a mapping between the assigned features $A(k)$ to coarse class $k$ and the $n$ fine prototypes associated to class $k$. Let $K_f= n*K$ be the total number of fine prototypes and $U$ the number of pixels in the training batch.  We first compute softmax scores from the similarities between 3D features and 3D fine prototypes $\{p^{3D}_{fk}\}_{k=1}^n$ and the similarity between 2D features and 2D fine prototypes $\{p^{2D}_{fk}\}_{k=1}^n$. 
Subsequently, for a coarse class $k$, we solve \cref{eq:ot} with $S^f \in \Re^{|A(k)|n}$ defined as $S^f_{ik} = exp(F^{2D}_i p^{2D}_{fk}/\tau_2) exp(F^{3D}_ip^{3D}_{fk}/\tau_2) / Z$ yielding $Q_k^f \in \Re^{|A(k)|n}$, Z being a normalization factor.
The $K$ resulting mappings $Q_k^f$ are combined into $Q^f \in \Re^{UK_{f}}$, which is used as segmentation target to optimize the overall training objective \cref{eq:ce}  with regard to the image encoder \IE{} and the segmentation/geometry fields \SF{}/\GF{}. 
Note that with this hierarchy constraint, $Q^{f}$ is not a minimizer of \cref{eq:ce}  for the fine feature/prototypes similarity softmax score but still provides coherent and valid target labels.
Fine assignments are defined as $A_f(k) = \{ i | h(i)=k\}$ with $h=argmax_k(Q^{f})$. 
The fine prototypes are updated by EMA based on fine assignments, similar to the update of the coarse prototypes.

We want to enforce, in the embedding space, the hierarchy that was defined through the aforementioned hierarchical clustering algorithm. To that end we introduce the following hierarchical prototype loss.
$$\textstyle{
L_{hierar} =  \frac{1}{2}  \sum_{k=1}^{K} \frac{1}{|A(k)|} \sum_{i \in A(k)} L_{ick} }$$ 
$$\textstyle{+ \sum_{k=1}^{K_f} \frac{1}{|A_f(k)|} \sum_{i \in A_f(k)} max(L_{ifk}, max_{j\in A(k^c)}(L_{jck^{c}}))}$$

where $L_{ick}$ and $L_{ifk}$ are the pixel to prototype contrastive losses at coarse ($c$) and fine level ($f$), with 
$$\textstyle{L_{ick} = -\log\left(\frac{\exp{(F_{u_i} p_{ck}}/\tau )}{ \sum_{j=1}^N \exp({F_{u_i} p_{ck}/\tau})}\right)}$$
and $L_{ifk}$ defined similarly.
For simplicity, we omitted the 2D/3D notations, but the loss is independently applied on 2D feature/prototypes and 3D feature/prototypes. Note that
the  $k^c$ coarse prototypes are associated to the fine prototypes $k$ and
minimizing these distances between pixels and their assigned prototypes ensures that the loss between each pixel and its corresponding fine prototype remains below the loss with the associated coarse prototype.

\PARR{Uncertainty} is  modeled  by adding an uncertainty prediction head so that the 2D encoder predicts both segmentation logits $l \in R^{KWH}$ and class-wise uncertainties $u \in R^{KWH}$ for each pixel.
In this formulation, the network predicts a per-pixel Gaussian distribution over the segmentation classes, where the vector of logits $l_i$ is the mean and the vector of uncertainties $u_i$ is the diagonal elements of the covariance matrix. 
The cross entropy objective in \cref{eq:ce} is therefore modified by sampling the Gaussian distribution before applying softmax. This  yields:
$$\textstyle{
 s^{2D}(k|u_i) = \softmax(\mean_t(\{l_i^{2D} + u_i *\epsilon_t\}_1^{N_S}))} \enspace ,$$ 
 $$\textstyle{
 s^{3D}(k|u_i)= \softmax(\mean_t(\{l_i^{3D} + u_i *\epsilon_t\}_1^{N_S}))} \enspace,$$ where $\epsilon_t \sim N(0,I)$ is jointly sampled $N_S$ times for the 2D and 3D logits so that both modalities benefit from the uncertainty modelling. 
 This allows for better stability during training and we use uncertainty to filter out ambiguous pixels during visual localization.

 We provide visual examples of coarse/fine image-based and rendered segmentations,  as well as coarse/fine uncertainty maps in \cref{fig:more_seg_visu}.

 \begin{algorithm*}
 \caption{Pseudo algorithm describing the training process of PPNeSF. }
  \label{alg:train}
      
 \SetAlgoLined
 \KwData{Set of posed training images with associated depth}
 Randomly initialize the coarse and fine prototypes.\\
 \For{iteration in range(N\_iterations)}{
    Sample a random image $I$ \\
    Extract 2D image-based features, fine/coarse segmentation maps and uncertainty $F^{2D},s^{2D}_c,s^{2D}_f,u_c,u_f$  \\
    Sample 4096 rays $\{r_n\}_{n=1}^{4096}$ in $I$, through two rounds of proposal networks sample 24 final samples per ray\\
    Bilinearly interpolate $F^{2D}$ at ray origin pixels which yields $\{F^{2D}_i\}_{n=1}^{4096}$ \\
    Query the feature field \FF{} and \PPNeSF's \GF{}-\SF{} to a obtain feature, opacity, coarse and fine segmentations per 3D point $f^{3D}_i,o_i,s_{ic},s_{if}$ \\
    Integrate along rays to obtain rendered feature, opacity and segmentation maps $F^{3D},d,s^{3D}_{c},s^{3D}_{f}$ \\
    Compute depth loss $L_{depth}$, distortion loss $L_{dist}$, anti inter-level loss $L_{interl}$ \\
    Compute feature contrastive loss $L_{NCE}$ (see \cref{eq:Lnce}) with $\{F^{3D}_i,F^{2D}_i\}$\\
    Solve \cref{eq:ot}
    to obtain coarse segmentation targets $Q$, find assignments per class A(k) \\
    Compute coarse cross entropy loss \cref{eq:ce} with coarse assignment $Q$ and coarse segmentations $s^{3D}_{c},s^{2D}_{c}$  \\
    \For{coarse class k in range(K)}{
        With fine prototypes associated to class k and feature assigned to class k, solve \cref{eq:ot} 
        to obtain partial fine segmentation targets $Q_kf$, find assignments per fine class $A_f(k)$ \\
    }
    Combine partial fine into fine assignments $Q_f$ and compute fine cross entropy loss  \cref{eq:ce} \\
    EMA update on coarse and fine prototypes based on coarse and fine assignments \\
    With coarse/fine assignments $A,A_f$, coarse/fine prototypes and $\{F^{3D}_i,F^{2D}_i\}$ compute $L_{hierar}$ \\
    
    Jointly update the feature field \FF,  segmentation-geometry fields \SF{}-\GF{},  encoder \IE{} by minimizing  $L = 2*L_{depth} + 0.5*L_{dist} + 0.1*L_{interl} + 0.2*L_{NCE} + 0.2*L_{CEc} + 0.2*L_{CEf} + 0.05*L_{hierar}$ 
    }

\end{algorithm*}

\section{Details on training and localization}
\label{sec:training}

\PAR{Training} \PPNeSF{} and the feature field \FF{}  are trained with an 
initial learning rate of 1e-2 exponentially decaying to 1e-4 with the Adam optimizer \cite{kingma2014adam}.
The encoder is trained with an initial learning rate of 1e-3 exponentially decaying to 1e-4 with the Adam optimizer \cite{kingma2014adam}. 

All models are trained for 50k iterations on a single Nvidia V100 32Gb GPU. 
We sample 4096 rays per training iteration. 
A training iteration takes approximately 800ms. 
50,000 training iterations are done per scene. Proposal weights are annealed during 1,000 iterations.
The coarse and fine prototypes are randomly initialized at the beginning of the training.
Similar to \cite{barron2023zip}, distortion (with a weight of 0.5) and anti-interlevel (with a weight of 0.1) losses are used during training. 

Depth supervision with the depth loss from \cite{yu2022monosdf} (with a weight of 2) is used to give coarse geometry priors to our models.  
We used  rendered depth maps from (\url{https://github.com/vislearn/dsacstar}) on 7Scenes, 
Marigold \cite{ke2023repurposing} (model "{\tt marigold-lcm-v1-0}") to estimate depth maps
on Cambridge Landmarks, 
and ZoeDepth \cite{bhat2023zoedepth} ("{\tt zoe nk}" model on torch hub) on Indoor6. 

To resume these details, we provide the pseudo-code describing the training process of PPNeSF in \cref{alg:train}.

\PAR{Visual localization}
During the pose refinement, 4096 rays are sampled per iteration on 7Scenes and Indoor6, while 8192 rays are sampled per iteration on Cambridge Landmarks. 
64 proposal samples per ray and 32 final samples per ray are used. 
The pose is refined for 150 coarse iterations and 150 fine iterations. 
While most of the queries converge in a much lower number of iterations, a few queries require more iterations when the optimization landscape is not smooth. 
The pose is optimized on SE(3). 
with an Adam optimizer, 0.33 decay rate  
where the initial learning rate is set to  2e-2 for all datasets. 

Comparatively, we run a simple baseline performing pose refinement by aligning RGB images rendered with ZipNeRF to the query images. It yields an average 19cm/4.8°/47\% on 7Scenes 
and 5.1m/31° on Cambridge Landmarks compared to 5cm/0.35°/87\% on 7Scenes 
and 14cm/0.4° on Cambridge Landmarks obtained with \PPNeSF. The poor localization results obtained with 
RGB alignment is mainly due to to illumination variations and lack of texture, which is compensated by the segmentation based representations of  \PPNeSF{}, which is robust to such challenges.

\PAR{Training/runtime cost \vs performance}
\PPNeSF{} provides higher localization accuracy and a better degree of privacy than concurrent methods thanks to its better representation learning/clustering approach and regularizations. But this 
comes at a higher training cost. 
In \cref{tab:trainingtime},  we report for each dataset the
average training time, as well as 
average inference time for an image.
Note that efficiency during visual localization could be improved through diverse optimization techniques, however working on this aspect was out of the scope of this paper.

\begin{table}[tt!]
\scriptsize
\centering
\resizebox{1.0\linewidth}{!}{
  \begin{tabular}{l||c|c|c|c}
     Runtimes  & 7Scenes & Indoor6 & 360 & CL \\
     \hline
     Training    & 9h23  & 9h05  & 8h38  & 10h44  \\
      Refinement   & 15s & 15s & X & 15s \\
    \end{tabular}
}
\caption{Average training and average per-image inference times per dataset.}
      \label{tab:trainingtime}
      \vspace{-.5cm}
    \end{table}

\begin{figure*}[tt!]
\centering
\includegraphics[width=0.85\linewidth]{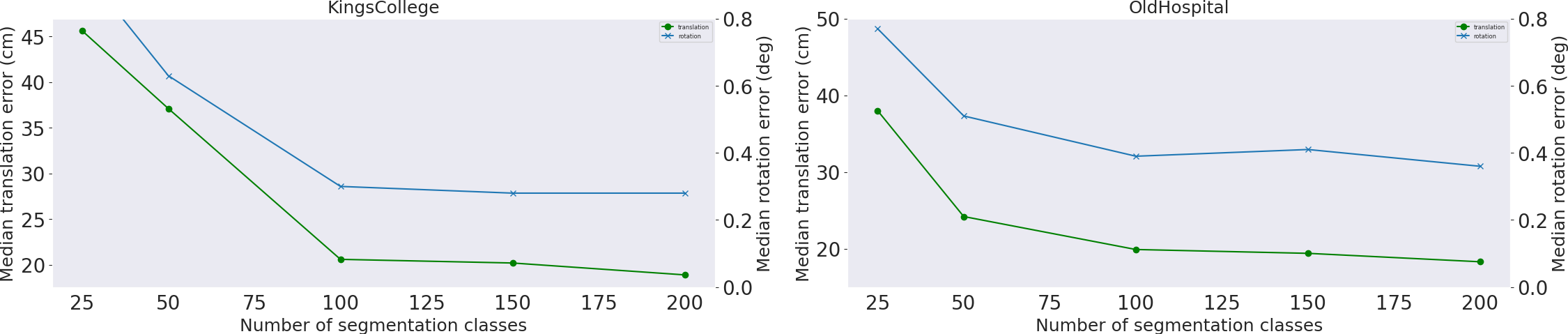}
\caption{Refinement with different number $K_f=n*K$ of fine classes (varying number $K$ of coarse classes, $n=5$ fine classes per coarse class). 
The green curve corresponds to the translation and the blue curve to the rotation error.
}
\label{fig:abl_nclasses}
\end{figure*}

\PAR{The impact of the number of classes}
To evaluate the impact of the numbers of latent segmentation classes, we train the model with different number of classes on the following two scenes: \emph{OldHospital} and \emph{KingsCollege} from the Cambridge Landmarks dataset. We show median translation errors and rotation errors against number of classes in \cref{fig:abl_nclasses}. 
On \emph{KingsCollege}, at least 100 classes are required to accurately refine the pose while on \emph{OldHospital} 50 classes are enough (\emph{OldHospital} is less complex than \emph{KingsCollege} so most information within the scene can be captured with fewer classes). Above these minimum number of classes, accuracy increases at a marginal rate.  For a small number of classes the localization accuracy is poor, meaning that a minimum number of classes is required to provide sufficient discriminativeness.
Overall, the optimal number of classes is scene-dependent and depends on the complexity of the scene.

\section{Evaluating privacy preservation}
\label{sec:more_privacy_exps}

\PAR{Architecture and optimization} We use a UNET~\cite{RonnebergerMICCAI15UNetConNetworks4BiomedicalImgSegm} architecture, which takes as input feature maps and outputs one channel grayscale reconstructed images. 
It consists of six encoding blocks (each composed of the following layers: Reflectionpad2D, conv2D, Batchnorm2D and ReLU activation), six decoding blocks (each composed of the following layers: Upsample, Reflectionpad2D, conv2D, Batchnorm2D and ReLU activation) of four refinement blocks (each composed of the following layers: Reflectionpad2D, conv2D, Batchnorm2D and ReLU activation or sigmoid activation and no batchnorm for the last block). 
Internal dimensions of encoder blocks are (256, 256, 256, 512, 512, 512), internal dimensions of decoder blocks are (512, 512, 512, 256, 256, 256), internal dimensions are refinement blocks are (128, 64, 32, 1), the encoder blocks conv2d kernel size is 4, the encoder blocks conv2d kernel size is 3. 
Skip connections are used for the decoder blocks.
The inversion model is trained with an Adam optimizer \cite{KingmaICLR15AdamStochasticOptimization}, the initial learning rate set to 1e-3 and weight decay set to 1e-4. 
It is trained for 50 epochs and each epoch containing 100 rendering iterations with a batch size of 2. 
The scene is changed (and the associated implicit model trained on the same scene) once every epoch and the learning rate decays once every epoch. 
The set of scenes used for training one inversion model belong to the same dataset.

\PAR{Training protocol} For each training epoch of the inversion model, we randomly select a scene along with a Neural Implicit Field (\NIF{}) trained on that scene (note that the \NIF{} is then frozen as we train only the inversion model).
Then for each iteration,  
we select a viewpoint with an associated image. 
On the rays emerging from this camera viewpoint, points are sampled and internal representations from the implicit model are extracted.
These representations are then rendered by alpha composition using the opacity weights of the implicit model.
The rendered internal representations are fed to the inversion network which 
attempts to reconstruct the grayscale GT image by minimizing a combination of L1 loss and perceptual LPIPS loss.
We learn to reconstruct the grayscale images -- instead of RGB -- to increase the generalization power of the model across datasets and to make the  model robust to color variations of certain objects.

\PAR{Evaluation protocol} We use the FID implementation of {\tt torcheval} metrics and the LPIPS implementation of \url{https://github.com/richzhang/PerceptualSimilarity.git}. 
These metrics are applied on grayscale images with 3 replicated channels. 
Overall the more visually similar are the reconstructed images to the original (grayscale) ones, the higher is the risk that the \NIF{} contain fine grained details, hence sensitive information about the scene, implying lower privacy.
We use the publicly available LLaVa mode "{\tt liuhaotian/llava-v1.5-7b}" to describe the grayscale original and reconstructed images. 
Max new tokens is set to 1024 and  num beams to 4. 
The KeyBert implementation used is from \url{https://github.com/MaartenGr/KeyBERT.git}  is used.

\subsection{Additional privacy baseline}
\label{sec:further_baselines}

\PAR{Training \PPNeSF{} with an additional photometric loss}
In addition to \ZipNeRFwoRGB{} and \PPNeSF{}, we add another baseline called \RGBPPNeSF{} in which we train \PPNeSF{} with an additional photometric loss. Note that, similar to \ZipNeRFwoRGB, we remove the color head of \RGBPPNeSF{} after training. Following the privacy attack and evaluation protocol from \cref{sec:privacy}, we train inversion models and evaluate the privacy of this new baseline on three datasets (7Scenes, Indoor6, Mip360). The reconstruction results are displayed in \cref{tab:full_recons_privacy}. 
Adding a photometric loss during training subsequently allows the inversion to recover more detailed and accurate images, indicating a degraded degree of privacy compared to \PPNeSF. 
This further validates our hypothesis according to which using RGB supervision induces a privacy breach. This conclusion can be extended to any Nerf method using RGB supervision as they use similar architectures and optimization processes as our baselines. This further confirms that traditional NeRF-based localization methods are not privacy preserving.

\begin{figure*}[tt!]
\centering
\includegraphics[width=1.\linewidth]{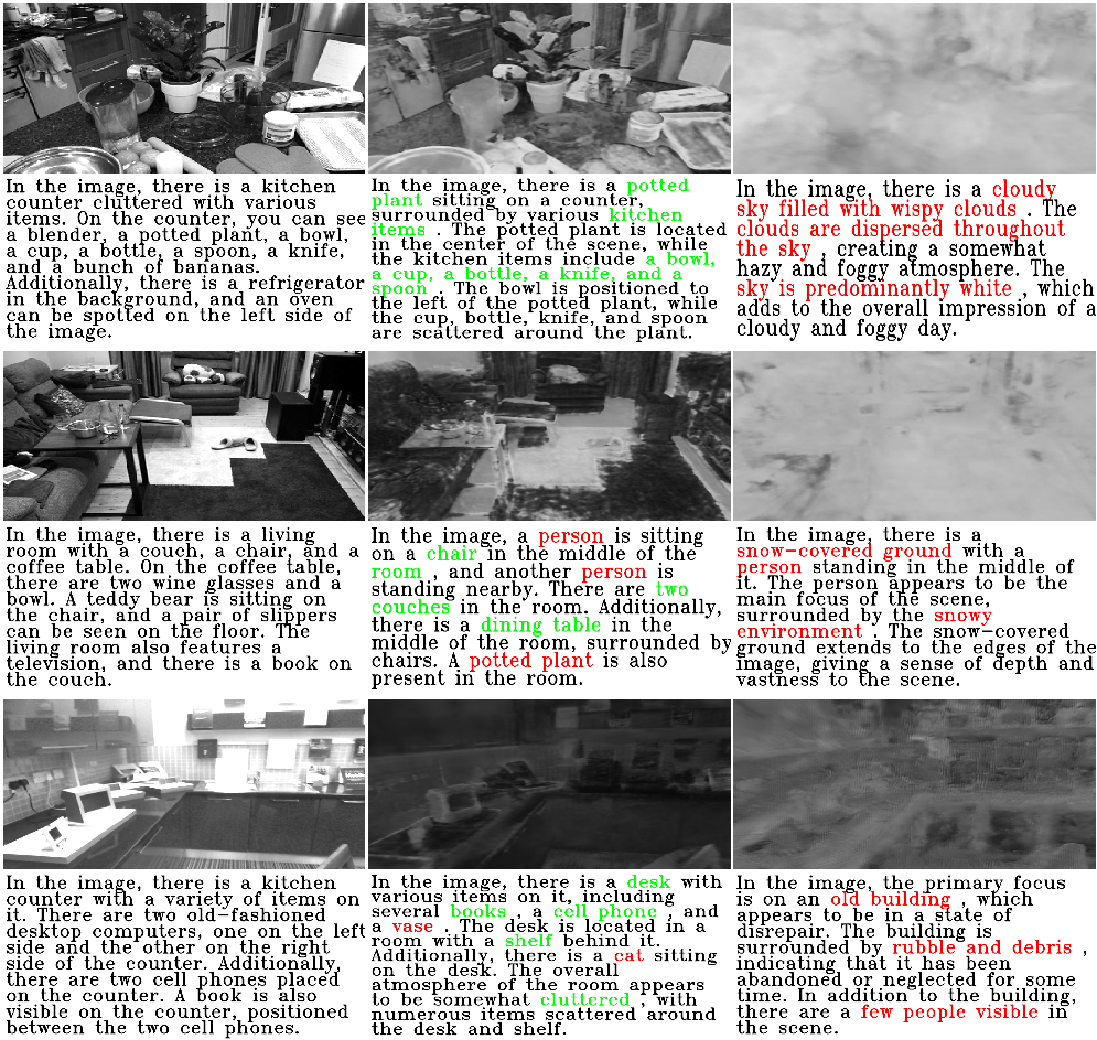}
\caption{From left to right: Ground truth image, image reconstructed from \ZipNeRFwoRGB, image reconstructed from \PPNeSF{} with the inversion attack. Below each image we show the associated LlaVa's descriptions highlighting in {\color{green}green} objects correctly identified by the VLM and
in {\color{red}red} hallucinated objects.}
\label{fig:more_privacy_visu}
\end{figure*}

In \cref{fig:more_privacy_visu} we also display additional comparisons between reconstructed images from \ZipNeRFwoRGB and \PPNeSF{}, along with the associated LLaVa description for each image in the figure with the prompt "Precisely list the objects and details which can be seen in the image". 
Our approach obfuscates most of the high frequency and texture details compared to models using RGB as supervision.

\begin{table*}[t!]
\begin{center}
\resizebox{0.85\linewidth}{!}{
  \begin{tabular}{c||c|c|c|c|c}
    Metric  & \ZipNeRFwoRGB & \ZipNeRFwoRGB{} 8bits & \RGBPPNeSF & \RGBPPNeSF{} 8 bits & \PPNeSF \\
  \hline
    LPIPS($\uparrow$) & 0.53 & 0.59 & 0.55 & 0.54 & \textbf{0.60} \\
    FID($\uparrow$) & 233 & 241 & 296 & 262 & \textbf{313} \\
  
     \hline
     \hline
    \end{tabular}
}
\end{center}
    \caption{\textbf{Comparison to additional privacy baselines.} The inversion model is trained on the mip360 dataset and evaluated on the Chess scene from the 7Scenes dataset. We evaluate image reconstruction quality through the LPIPS($\uparrow$)/ FID ($\uparrow$) metrics. Higher LPIPS/FID on images reconstructed from our privacy attack indicates a more privacy-preserving neural field model.
}
      \label{tab:8bit_privacy}
    \end{table*}

\begin{table*}[tt!]
\tiny
\resizebox{\linewidth}{!}{
  \begin{tabular}{cl||c|c|c|c|c|c|c|c}
   &  Model  & Chess & Fire & Heads & Office & Pumpkin & Redkitchen & Stairs  & Average\\
    \hline
   \multirow{3}*{\begin{sideways} Tr360 \end{sideways}}  
    & \ZipNeRFwoRGB &  0.53/233 & 0.55/347 & 0.56/323 & 0.55/230 & 0.54/217 & 0.57/232 & 0.56/173 &  0.55/250\\ 
   & \RGBPPNeSF & 0.55/296 & 0.59/405 & 0.57/337 & 0.57/321 & 0.51/217 & 0.55/260 & 0.51/193 & 0.54/289 \\
   & \PPNeSF &  0.60/313 & 0.62/425 & 0.60/389 & 0.62/282 & 0.58/284 & 0.60/304 & 0.57/257 & \textbf{0.59/322}\\
   \hline
    &    & Bicycle & Bonsai & Counter & Garden & Kitchen & Room & Stump  & Average\\
\hline
 \multirow{3}*{\begin{sideways} Tr7S \end{sideways}}  
 & \ZipNeRFwoRGB & 0.69/321 & 0.49/193 & 0.53/240 & 0.64/150 & 0.62/321 & 0.57/301 & 0.66/328 & 0.6/264\\
   &\RGBPPNeSF & 0.72/241 & 0.63/362 & 0.61/391 & 0.70/269 & 0.62/306 & 0.60/284 & 0.73/413 & 0.65/323 \\
  &\PPNeSF &  0.81/297 & 0.71/446 & 0.72/470 & 0.81/308 & 0.74/381 & 0.69/386 & 0.81/448 & \textbf{0.76/390} \\
    \hline
   &    & scene1 & scene2a & scene3 & scene4a & scene5 & scene6 &   & Average\\
    \hline
     \multirow{3}*{\begin{sideways} Tr 7S \end{sideways}}  
    & \ZipNeRFwoRGB & 0.51/262 & 0.53/372 & 0.52/334 & 0.55/267 & 0.50/253 & 0.53/302 & & 0.52/298 \\
    &\RGBPPNeSF & 0.52/245 & 0.57/292 & 0.56/268 & 0.60/275 & 0.50/208 & 0.56/0.53/303 & & 0.55/265 \\
    & \PPNeSF & 0.56/311 & 0.61/364 & 0.57/324 & 0.61/334 & 0.56/279 & 0.57/316 & & \textbf{0.58/321} \\
\hline
     \hline
    \end{tabular}
}
\caption{We evaluate image reconstruction quality through the LPIPS($\uparrow$)/ FID ($\uparrow$). We train an inversion model on one dataset (7S: Tr7S, 360: Tr360) and reconstruct images from another unseen dataset. Note that lower reconstruction quality implies better privacy.}
      \label{tab:full_recons_privacy}
    \end{table*}

\begin{figure*}[tt!]
\centering
\includegraphics[width=0.75\linewidth]{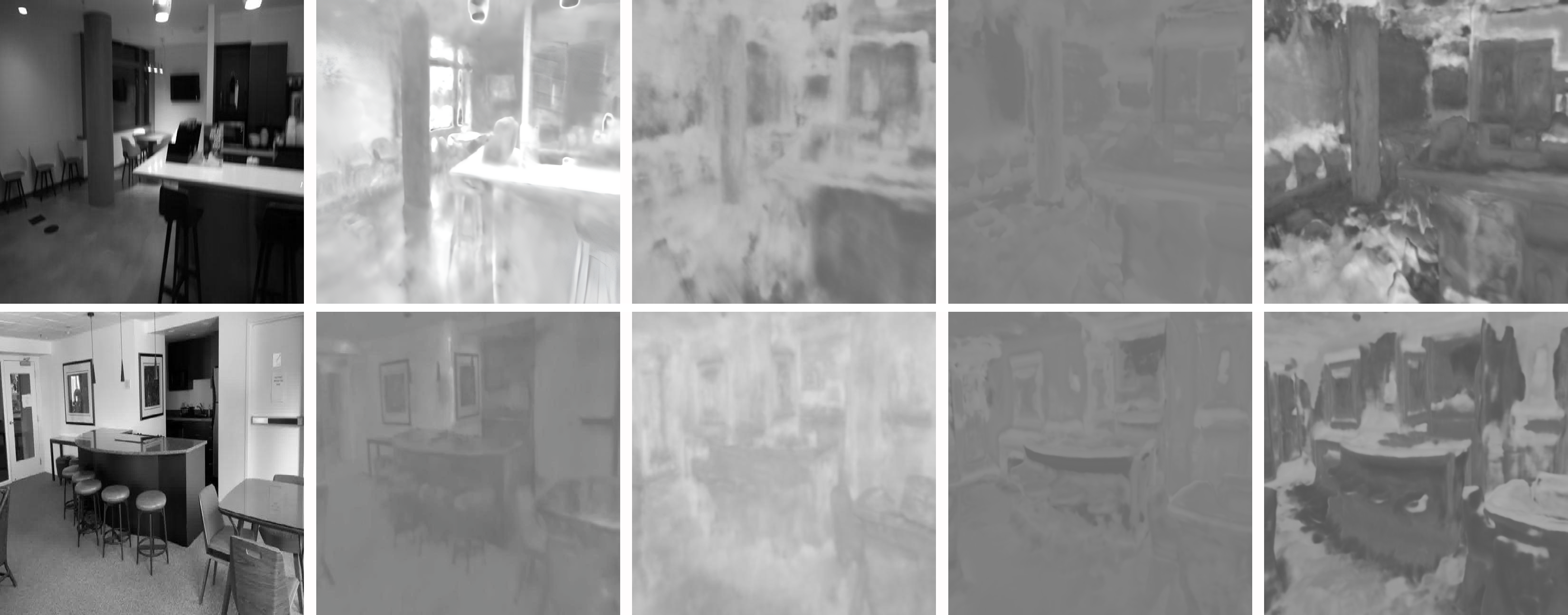}
\caption{\textbf{Left to right}: GT image, images reconstructed 
from \ZipNeRFwoRGB / \PPNeSF / Q-Segs / \PPNeSF+R-Segs. Segmentations provide little additional information.} 
\label{fig:inv_segs}
\end{figure*}

\begin{table*}[tt!]
\begin{center}
\resizebox{0.85\linewidth}{!}{
  \begin{tabular}{l||c|c|c|c}
    &  \multicolumn{4}{c}{LPIPS($\uparrow$)/ FID ($\uparrow$) / Captions similarity ($\downarrow$)} \\
     Model  & \ZipNeRFwoRGB & PPNeSF & Q-Segs & PPNeSF+R-Segs \\
     \hline
     Metrics & 0.52 / 298 / 0.55 & \textbf{0.58} / \underline{321} / \underline{0.43} & 0.55 / \textbf{324} / \textbf{0.42} & \underline{0.57}/ 306 / 0.48\\
    \end{tabular}
    }
    \end{center}
\caption{ \textbf{Privacy of the segmentations.}
 Indoor6 inversions results (trained on 7Scenes). Higher LPIPS/FID and lower captions similarity implies better privacy.}
      \label{tab:inv_segs}
    \end{table*}

\PAR{Privacy experiments with 8bit images}
In this section we explore an alternative option where instead of training NeRF models with continuous RGB images, we use discretized 8bit RGB images containing obviously much less fine-grained information. On the \textit{Chess} scene (7Scenes dataset), we re-train our privacy baselines (\ZipNeRFwoRGB{} and \RGBPPNeSF) using 8bit RGB images instead of the original RGB images. On the mip360 dataset, we train our privacy baselines (\ZipNeRFwoRGB{} and \RGBPPNeSF) using 8bit RGB images and train inversion models (see \cref{sec:privacy} for more details) to recover images on the same mip360 dataset. Finally, we apply the inversion model on the privacy baselines trained on \textit{Chess} and evaluate the quality of the reconstructed images in \cref{tab:8bit_privacy}. Replacing RGB with 8bit  images slightly increases the level of privacy, but it remains much lower than \PPNeSF{}, which does not use image level supervision. Indeed, 8bit color images still contain much more information than our segmentation maps that are based on 100 
scene specific clusters. 
It also shows that training neural implicit fields with discretized information only still yields a coherent and accurate neural field.

\subsection{Privacy of the segmentations}
\label{sec:segmprivacy}

This paper tackles the privacy of NeRF models by analysing the information contained in the internal space of the neural fields 
and accordingly designing a solution to remove privacy sensitive content 
from these fields.
Regarding the privacy of segmentations (either rendered or extracted images), we refer to \cite{pietrantoni2023segloc}, which shows that segmentations increase privacy by preventing inversion through non-injective RGB→classes mappings. 
Note that contrarily to \cite{pietrantoni2023segloc}, our segmentations are trained per scene and correspond to a quantization of PPNeSF’s already privacy-preserving internal features which further increases the degree of privacy.
To illustrate this, we train, on 7Scenes, inversion models taking as input query segmentation maps (Q-Segs), rendered segmentation maps (R-Segs) and rendered segmentation maps combined with internal ppNeSF features. For each modality we then evaluate the inversion model on Indoor6. We report LPIPS/FID and captions similarity between the original and recovered images in \cref{tab:inv_segs}. We observe that  segmentation still provide a much higher degree of privacy than the privacy baseline \ZipNeRFwoRGB. Adding rendered segmentations to internal \PPNeSF{} features only degrades very slightly the level of privacy,  which further validates our assumption about the privacy of segmentations. Example visualizations are provided in \cref{fig:inv_segs}.

\end{document}